%% file: cvpr24_yenyo_arxiv_cr.tex
\documentclass[10pt,twocolumn,letterpaper]{article}

\usepackage[pagenumbers]{cvpr} 

\input{preamble}

\usepackage{multirow}
\usepackage{multicol}
\usepackage{color}
\usepackage{colortbl}
\usepackage{amsmath,amssymb}
\usepackage{textcomp}
\usepackage{tikzinclude}
\usepackage[pdftex]{preview}
\usepackage[accsupp]{axessibility} 

\definecolor{cvprblue}{rgb}{0.21,0.49,0.74}
\usepackage[pagebackref,breaklinks,colorlinks,citecolor=cvprblue]{hyperref}

\def\paperID{7424} 
\def\confName{CVPR}
\def\confYear{2024}

\title{Diffusion Reflectance Map:\\%
Single-Image Stochastic Inverse Rendering of Illumination and Reflectance}

\author{Yuto Enyo \qquad Ko Nishino\\
Graduate School of Informatics, Kyoto University\\
{\tt\small \url{https://vision.ist.i.kyoto-u.ac.jp/}}
}

\begin{document}

\twocolumn[{
\maketitle
\begin{center}
    \vspace{-5mm}
    \captionsetup{type=figure}
        \includegraphics[width=1.\linewidth]{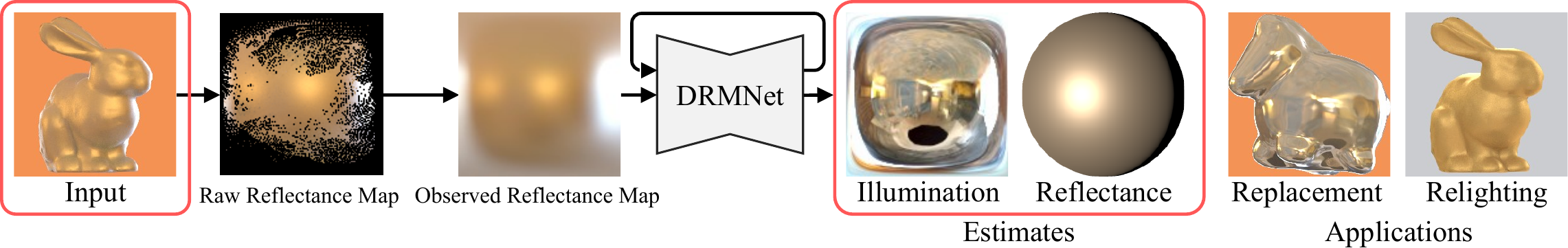}
        \captionof{figure}{We introduce the first single-image stochastic inverse rendering method, a principled approach for recovering the attenuated frequency spectrum of the illumination and reflectance by seamlessly integrating a neural generative process in inverse rendering. Our key idea is to recover the illumination as a reflectance map of a perfect mirror. The results enable arbitrary object insertion and relighting.}
    \label{fig:opening}
\end{center}
}]

\begin{abstract}

    Reflectance bounds the frequency spectrum of illumination in the object appearance. In this paper, we introduce the first \textit{stochastic} inverse rendering method, which recovers the attenuated frequency spectrum of an illumination jointly with the reflectance of an object of known geometry from a single image. Our key idea is to solve this blind inverse problem in the reflectance map, an appearance representation invariant to the underlying geometry, by learning to reverse the image formation with a novel diffusion model which we refer to as the Diffusion Reflectance Map Network (DRMNet). Given an observed reflectance map converted and completed from the single input image, DRMNet generates a reflectance map corresponding to a perfect mirror sphere while jointly estimating the reflectance. The forward process can be understood as gradually filtering a natural illumination with lower and lower frequency reflectance and additive Gaussian noise. DRMNet learns to invert this process with two subnetworks, IllNet and RefNet, which work in concert towards this joint estimation. The network is trained on an extensive synthetic dataset and is demonstrated to generalize to real images, showing state-of-the-art accuracy on established datasets.

\end{abstract}

\section{Introduction}

Light interacting with surfaces narrate our rich visual world. The light from illuminants reflect off object surfaces deeply entangling the surface material and geometry in its process. Deciphering the surface geometry, reflectance properties, and the incident illumination from this reflected radiance has been the central focus of physics-based vision research. Our focus is achieving this from a single image.




The ability to explicitly recover rich situational information from a single glance at an object goes to the very heart of embodied perception and enables effective use of visual information both in human perception and robot vision. For instance, decisions on how to act on an object needs to be made as the visual information is captured, and cannot be stalled until a hundred images are captured. 

Even when the shape is known, \ie, with an RGB-D smartphone sensor, disentangling the reflectance and illumination from a single image remains ill-posed due to two key problems. The first is color ambiguity---it is unclear whether to associate, for instance, the green object appearance to the illumination or the reflectance which cannot be resolved without assumptions on the color of either. The bigger and fatal obstacle is the loss of information. The reflectance acts as an angular filter on the incident illumination while the geometry determines its orientation \cite{ramamoorthi2001signal}. As a result, the frequency spectrum of the object appearance is bounded by the highest frequency of either the reflectance or the illumination, the former of which is usually lower \cite{basri2003lambertian}. 

The frequency bound limits the utility of the estimated reflectance and illumination of past methods which do not pay attention to it. Due to low-pass reflectance, the illumination estimate is blurry and can only be used to insert an object in the scene that has lower frequency characteristics than the object used to recover the illumination. The same can be said about relighting using the illumination estimate. 

How should we recover the true illumination and reflectance from object appearance? The frequency bound of the object appearance suggests that there is no such thing as a true estimate. Given the object appearance, the illumination must be generated together with the reflectance as only its combined, partial information is encoded in it. Inverse rendering, particularly from a single image, as such, is a generative process. We argue for \textit{stochastic inverse rendering} in which image formation is explicitly modeled as a stochastic process and its inversion becomes a stochastic generative reverse process. Stochasticity needs to be seamlessly integrated into the radiometric disentanglement. 


In this paper, we derive a principled stochastic inverse rendering method for a single image. We assume homogeneous reflectance, which is the most challenging in terms of lacking visual information and serves as a foundation for multiple materials and objects. Our key idea is to learn to generate the illumination from the object appearance with a diffusion model on the reflectance map. By formulating the task on a geometry-invariant reflectance map, it enables radiometric disentanglement in the same domain eliminating the need for complex differentiable rendering.

Our problem is inherently blind, \ie, the reflectance acting as the forward operator is also unknown. We introduce a novel diffusion model, Diffusion Reflectance Map Network (DRMNet), that generates a reflectance map corresponding to a perfect mirror sphere from the observed reflectance map while jointly estimating the reflectance. 
DRMNet consists of two subnetworks to learn to invert the forward radiometric image formation process with Gaussian noise.

The first, IllNet, is a U-Net that stochastically converts the current reflectance map into that of a sharper reflectance until it becomes perfectly mirror. The second, RefNet, is a simple network that takes in the observed and current reflectance maps and the current time-step and estimates the object reflectance as a parametric model. RefNet, in concert with IllNet, learns to iteratively refine the reflectance estimate. We introduce another diffusion model to transform the single input image into a reflectance map for DRMNet input by completing missing regions and resolving the many-to-one mapping from pixels to surface normals.  


We train these diffusion models with a large synthetic dataset of reflectance maps of known reflectance and complex natural illumination and evaluate the accuracy of our method on a number of real and synthetic data. Through extensive comparisons with relevant works, we show that our method achieves state-of-the-art accuracy on this challenging task. We believe this explicitly generative approach to single-image inverse rendering would benefit a wide range of downstream applications not just for image synthesis but also for situational awareness and embodied interactions with the objects and surroundings.

\section{Related works}
\label{sec:related}


\paragraph{Inverse Rendering}


The disentanglement of object appearance into its radiometric constituents, namely the geometry, reflectance, and illumination, may inform about the object and the surroundings to a wide range of applications in computer vision, graphics, and robotics. This task is often referred to as inverse rendering \cite{marschner1998inverse}, radiometric decomposition \cite{Lombardi3DV16}, appearance modeling \cite{appmodelcourse}, and inverse optics \cite{Zygmunt01}. 

Many recent methods model this entanglement solely for novel view image synthesis, a representative approach of which is Neural Radiance Fields (NeRF) \cite{mildenhall2021nerf}. The results are impressive, demonstrating high-fidelity novel view synthesis of intricate objects. These methods, however, inherently rely on many multi-view ray samples of the scene \cite{goesele2006mvs,schoenberger2016sfm,mildenhall2021nerf,barron2021mip,yariv2020idr,verbin2021refnerf,dai2023hybrid,liu2023nero} or many images under different scene conditions captured from the same viewpoint \cite{woodham1980photometric,li2018learning,li2023inverse,li2023dani}, typically on the order of tens of images. In sharp contrast, our focus is on recovering as much as possible about the object and scene from a single observation by leveraging the key physical characteristics of the interaction of light with the object surface.



Single-image inverse rendering is severely ill-posed even when the object shape is known. Lombardi \etal \cite{lombardi2012reflectance} introduce an MAP estimation framework with analytical reflectance and illumination priors. Chen \etal \cite{chen2022invertible} model reflectance and illumination using deep neural networks and estimate them through optimization with differentiable rendering. Meka \etal \cite{meka2018lime} estimate the reflectance parameters with an image-space U-Net to convert the object appearance into a mirror and recovers illumination as a linear combination of low-order spherical harmonics estimated from diffuse shading and mirror reflection. This results in patched-up frequency spectrum as the mirror reflection only gives sparse high frequency details and the spherical harmonics only low frequency components. Our method achieves joint estimation of the reflectance and illumination in a space invariant to the surface geometry, \ie, reflectance map. A few past works estimate a reflectance map \cite{rematas2016deep} or an appearance map \cite{maximov2019deep}. These methods, unlike ours, do not disentangle the illumination and reflectance. 

Georgoulis \etal \cite{georgoulis2016deLightdet} regress the reflectance parameters and illumination as an environment map from a reflectance map. This method, however, does not account for the frequency attenuation in the observation and only recovers an arbitrarily bounded illumination. A few recent methods learn to estimate shape, reflectance, and illumination from a single image just for one class of objects (\eg, car) \cite{georgoulis2018reflectance, wu2021rendering} or by using supervision of synthetic data \cite{bi2020sirnet,wimbauer2022derendering}. Self-supervised methods use differentiable rendering to decompose single images into shape, reflectance, and illumination \cite{janner2017self,yu2018inverserendernet,chen2019learning,chen2021dibrpp,yu2021outdoor,yi2023weakly}. Most of these methods are limited to simple or low-frequency illumination. Even with a fully-supervised model, directly estimating a single high-frequency solution from observations that only contain low-frequency components will lead to an averaged low-frequency estimate, \ie, it cannot return a full spectrum estimate. Our focus is to break free of this fundamental limitation of inverse rendering by formulating it as a stochastic generative process. 


\vspace{-8pt}
\paragraph{Illumination Estimation}
Past methods have also tackled estimation of illumination alone \cite{georgoulis2017around,weber2018learning,park2020physicallyinspired,wei2020objectbased}. Yu \etal \cite{yu2023accidental} estimated the lighting from shiny objects whose shape, reflectance, and texture are known. Our method does not make assumptions on the object reflectance and jointly estimates it. Swedish \etal \cite{swedish2021objects} achieves high-frequency illumination estimation by leveraging cast shadows. Our method does not rely on high-frequency illumination that casts sharp shadows, \ie, the illumination can have arbitrary frequency characteristics.

Recent illumination estimation methods extrapolate the full surrounding from a partial scene captured in the limited single view frustum \cite{legendre2019deeplight,gardner2017learning,gardner2019deep,zhan2021sparse,zhan2021emlight,zhan2022gmlight,dastjerdi2023everlight}. This has also been demonstrated for spatially varying illumination in 2D \cite{song2019neuralillumination,garon2019fast,tang2022estimating,bai2023local} or 3D \cite{srinivasan2020lighthouse,bai2023deep}. Several methods estimate the geometry as depth or normals, together with the reflectance and illumination from a single scene image \cite{li2019inverse,wang2021learning,zhu2022learning,li2022physically,zhu2022irisformer}. These methods, however, fundamentally assume rich scene information to be present in the image, not just a single object, and high-frequency lighting estimation is conducted mainly through spatial interpolation.

In a concurrent work, Lyu \etal \cite{lyu2023dpi} propose a generative model of environment maps by applying probabilistic diffusion as a prior on the illumination in a conventional Bayesian framework \cite{lombardi2012reflectance,chen2022invertible}. The method samples multiple illuminations using this diffusion model. In contrast, our approach reformulates inverse rendering from the ground up as a stochastic inversion process by seamlessly integrating probabilistic diffusion as a forward radiometric formation model on the reflectance map. This eliminates the need of differentiable rendering to bridge the complex domain gap between the image and 3D object surface and enables stochastic generation of the illumination without separate explicit sampling. We show that our method achieves significantly higher accuracy.

\vspace{-8pt}
\paragraph{Diffusion for Inverse Problems}
 Denoising Diffusion Probabilistic Models \cite{Sohl15,Ho20} have been applied to inverse problems in computer vision including image deblurring, inpainting, and super-resolution \cite{kawar2022denoising, chung2023diffusion, chung2022improving, kawar2021snips}. These methods assume known, deterministic forward operators and modify the reverse process based on score-matching. Other methods have used diffusion models by conditioning on the given image \cite{ren2023multiscale, saharia2022palette,lugmayr2022repaint}. Chung \etal \cite{chung2023parallel} extend these to noisy non-linear inverse problems. 


The forward diffusion process can be extended to general signal degradation \cite{daras2023soft, bansal2022cold,Lee2022Progressive,hoogeboom2023blurring}. Rissanen \etal introduced ``heat dissipation'' as a forward process, and gradually generated low- to high-frequencies in its reverse process \cite{rissanen2023generative}. The forward operators in these methods are known. Our problem is similar but significantly more challenging as the operator, \ie, the object reflectance, is unknown---we have a blind inverse problem.

\section{Diffusion Reflectance Map}
\label{sec:drm}

We start by reformulating radiometric image formulation as a forward stochastic process on the reflectance map.

\subsection{Forward Reflectance Maps}

The reflected radiance $L_r$ at a surface point $\mathbf{x}$ due to incident directional illumination $L_i$ can be described with the rendering equation \cite{Kajiya86}
\begin{equation}
    L_r(\mathbf{x}, \omega'_o) =  \int_{\Omega'} f_r(\omega'_i, \omega'_o) L_i(\omega'_i)(\omega'_i\cdot\mathbf{n}(\mathbf{x})) d\omega'_i\,,
    \label{eq:rendering}
\end{equation}
where $\omega'_i$ and $\omega'_o$ are the incident and outgoing directions of light in the local coordinate frame (\ie, with $\mathbf{n}(\mathbf{x})$ as the north pole) of the surface point, respectively, $\Omega'$ is the upper local hemisphere of the surface point, and $f_r$ is the bidirectional reflectance distribution function (BRDF). We assume a surface with homogeneous material properties $f_r(\mathbf{x}, \omega'_i, \omega'_o) = f_r(\omega'_i, \omega'_o)$. Spatially varying BRDFs are often handled by pre-segmenting the object surface into regions of different BRDF, or by considering soft assignment indicator values, which we will consider in future work.

Estimation of the reflectance (BRDF $f_r$) and the illumination ($L_i$) from a single image ($L_r$ at each point in the viewing direction) is challenging even when the surface geometry ($n$ for every $\mathbf{x}$) is known and the reflectance is homogeneous. The key difficulty stems from the multiplication of the two radiometric ingredients and the integration over the upper hemisphere. These cause frequency attenuation and implicit angular coordinate transform that hamper reflectance and illumination recovery from the object appearance, which past methods tackled with complex non-linear optimization or approximate differentiable rendering. 

We can explicitly rotate each local direction into the global coordinate frame and approximate each local integration domain with the hemisphere centered around the viewing direction. This means that, if the surface point $\mathbf{x}$ can be uniquely determined by its surface normal, the surface radiance (\ie, object appearance) can be uniquely defined by $L_r(\mathbf{n})$, the reflectance map \cite{HornBook}
\begin{equation}
    L_r(\mathbf{n}) = \int_{\Omega} f_r(\omega_i(\mathbf{n}), \omega_o)L_i(\omega_i(\mathbf{n}))(\omega_i(\mathbf{n})\cdot\mathbf{n})d\omega_i\,.
    \label{eq:refmap}
\end{equation}
Different surface points with the same surface normal can have different radiance due to global light transport, such as cast shadows and interreflection. We assume that these variations can be resolved when we map the object appearance (image) to the reflectance map ($L_r(\mathbf{n})$) (\cref{sec:obsref}).

\begin{figure}[t]
    \centering
    \includegraphics[width=0.95\linewidth]{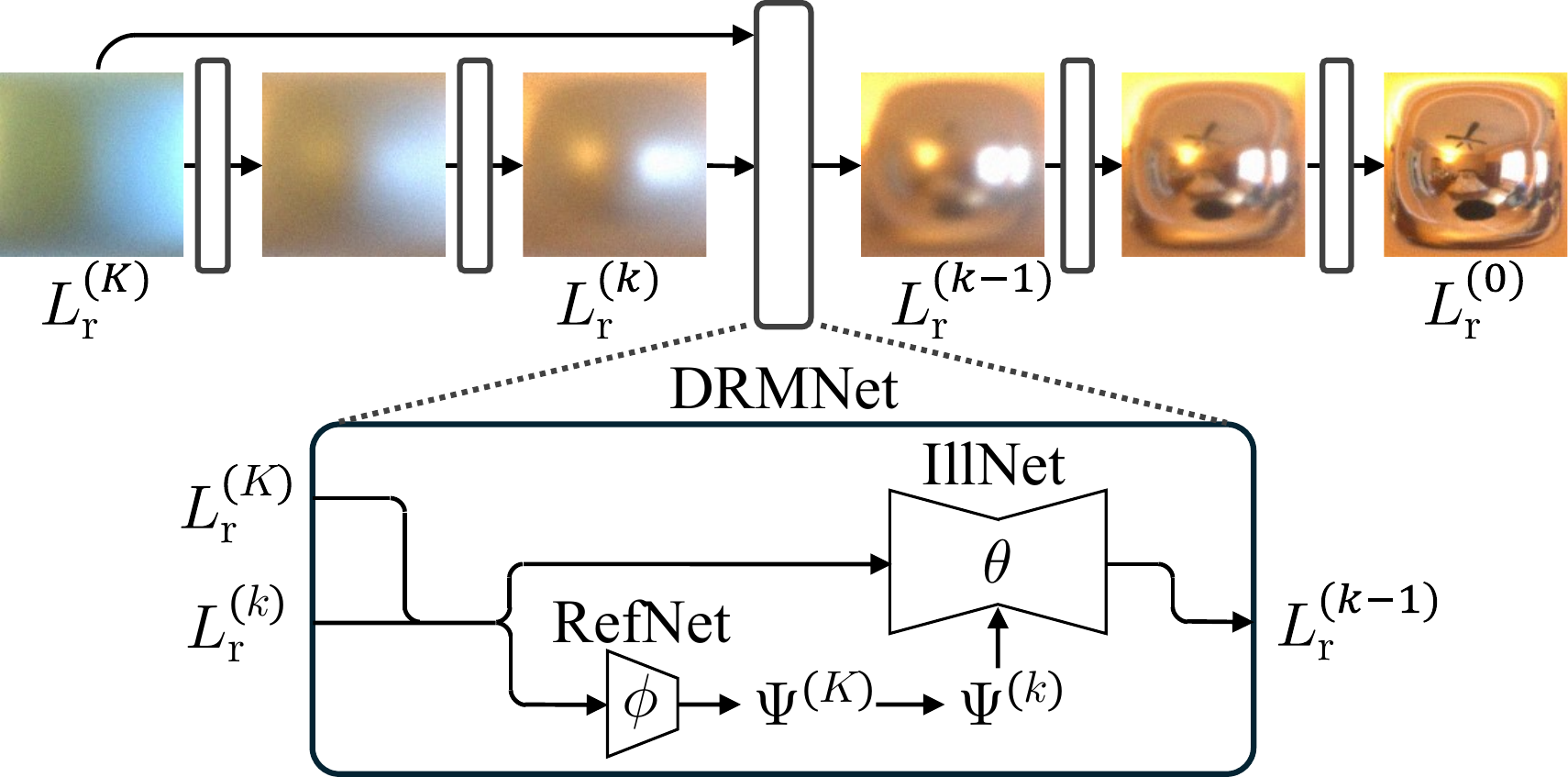}
    \caption{Overall architecture of Diffusion Reflectance Map Network (DRMNet). DRMNet consists of two subnetworks, IllNet for stochastic reverse diffusion to recursively transform the observed reflectance map into a reflectance map of a perfect mirror, and RefNet for jointly and iteratively estimating the reflectance.}
    \label{fig:drmnet}
\end{figure}
\subsection{Stochastic Inverse Rendering}
\label{sec:stochdis}

Optical image formation adds measurement noise which can be approximated with a zero-mean Gaussian. As such, image formation can be viewed as a Gaussian stochastic forward process on the reflectance map \cref{eq:refmap}. Now our goal is to recover the reflectance $f_r$ and the illumination $L_i$ in the global coordinate frame, \ie, as an environment map. This clearly necessitates a stochastic generative approach that reverts Gaussian added reflectance map to recover the attenuated frequency components of $L_i$ by $f_r$. 

The reflectance map enables us to formulate single-image reflectance and illumination disentanglement as a geometry-independent stochastic inversion process. This allows us to derive a stochastic inverse rendering approach, a principaled inverse rendering approach with stochasticity seamlessly integrated in the estimation process. In particular, we formulate it as generating a reflectance map of a perfect mirror surface from the observed image. This inverse diffusion process is not that of a regular noise to signal, but of a deterministic forward process with stochastic observation noise, similar to heat dissipation \cite{rissanen2023generative}. 

The inverse generative process can be considered as steps taken along the reflectance space towards a perfect mirror. If we employ an analytical reflectance model, this can be considered as a trajectory in its parameter space. We employ the Disney principled BSDF model \cite{Burley12}
\begin{equation}
\begin{split}
    &f_r(\boldsymbol\omega'_\mathrm{i},\boldsymbol\omega'_\mathrm{o};\Psi:=\{\boldsymbol{\rho}_\mathrm{d}, \rho_\mathrm{s}, \alpha, \gamma\})\\
    &= (1-\gamma)\frac{\boldsymbol{\rho}_\mathrm{d}}{\pi}(f_\mathrm{diff}(\boldsymbol\omega'_\mathrm{i},\boldsymbol\omega'_\mathrm{o})+ f_\mathrm{retro}(\omega'_\mathrm{i},\omega'_\mathrm{o};\alpha))\\
    & + f_\mathrm{spec}(\boldsymbol\omega'_\mathrm{i},\boldsymbol\omega'_\mathrm{o};\boldsymbol{\rho}_\mathrm{d}, \rho_\mathrm{s}, \alpha, \gamma)\,.
    \label{eq:disney}
\end{split}
\end{equation}
where we use a partial set of its BSDF model parameters for simplicity: $\boldsymbol{\rho}_\mathrm{d}, \rho_\mathrm{s}, \alpha$ and $\gamma$, the diffuse color, specular strength, roughness and metallicness, respectively. 
All of the parameters are normalized to $[0, 1]$. 
Please see the supplementary material for the full model. 

\subsection{Diffusion Reflectance Map Network}
\label{ref:drmnet}

\Cref{fig:drmnet} depicts the architecture of Diffusion Reflectance Map Network (DRMNet) which consists of two subnetworks. The first is a diffusion model, IllNet, that learns to invert the forward reflectance map generation process conditioned on the reflectance. The second is a reflectance estimator, RefNet. Input to DRMNet are the observed and current reflectance maps, $L_r^{(K)}$ and $L_r^{(k)}$, respectively. 


IllNet estimates the illumination from the observed reflectance map by iteratively generating a reflectance map $L_r^{(k-1)}$ and reflectance parameters one step closer to perfect mirror $f_r^{(k-1)}(\omega'_i,\omega'_o;\Psi^{(k-1)})$ from the current reflectance map $L_r^{(k)}$ and reflectance parameter estimate $\Psi^{(k)}$. The reflectance estimate of the object is $\Psi^{(K)}$, where $K$ is the maximum time step, \ie, the observation. This recursion traces a trajectory in the reflectance parameter space 
\begin{equation}
    \Psi^{(k - 1)} - \Psi_0 = \eta (\Psi^{(k)} - \Psi_0)\,,
    \label{eq:reverse-step-base-definition}
\end{equation}
where $\eta$ is a constant in $(0, 1)$ that controls the speed towards $\Psi_0$, \ie, a perfect mirror reflectance $[1, 1, 1, 1, 0, 1]$. 

The main parameter that determines the highest frequency of the reflectance is the surface roughness $\alpha$. When $\alpha=0$, the surface is perfect mirror reflectance, which we would like to achieve as $\Psi_0$. 
If we focus on the surface roughness $\alpha$, the iterative reverse steps defined in \cref{eq:reverse-step-base-definition} gradually moves it towards $0$ and its change 
    $\left| \alpha^{(k-1)} - \alpha^{(k)} \right| = (1-\eta)\alpha^{(k)}$,
asymptotically dampens in the process. As these smaller steps in the surface roughness correspond to higher frequencies in the reflectance map, this means, with the uniform steps in \cref{eq:reverse-step-base-definition}, the higher the frequencies, the denser the sampled steps. This contributes to robust recovery of the illumination.

This iterative generation, learned as a reverse diffusion process, starts with the observed reflectance map $L_r^{(K)}$ and reflectance parameters $\Psi^{(K)}$ and IllNet is applied recursively until it reaches a reflectance map $L_r^{(0)}$ corresponding to a perfect mirror $\Psi_0$. In practice, by definition (\cref{eq:reverse-step-base-definition}), $\Psi^{(0)}$ only asymptotically reaches $\Psi_0$. For this, we set $K$ to satisfy $||\Psi^{(0)} - \Psi_0||_2 < \epsilon$ for a small $\epsilon$. At inference time, we do not need to explicitly define $K$ and instead iterate till $\Psi^{(k)}$ reaches $||\Psi^{(k)} - \Psi_0||_2 < \epsilon$.

RefNet learns to estimate the reflectance parameters of the object in the observed single image: $\Psi^{(K)}$. The reflectance parameter $\Psi^{(k)}$ corresponding to the current reflectance map can be computed from this observed reflectance estimate by
\begin{equation}
    \Psi^{(k)} = \eta^{(K - k)}(\Psi^{(K)} - \Psi_0) + \Psi_0\,,
    \label{eq:refparam-def}
\end{equation}
using the steps taken so far $(K-k)$.

\begin{figure}[t]
    \centering
    \includegraphics[width=.9\linewidth]{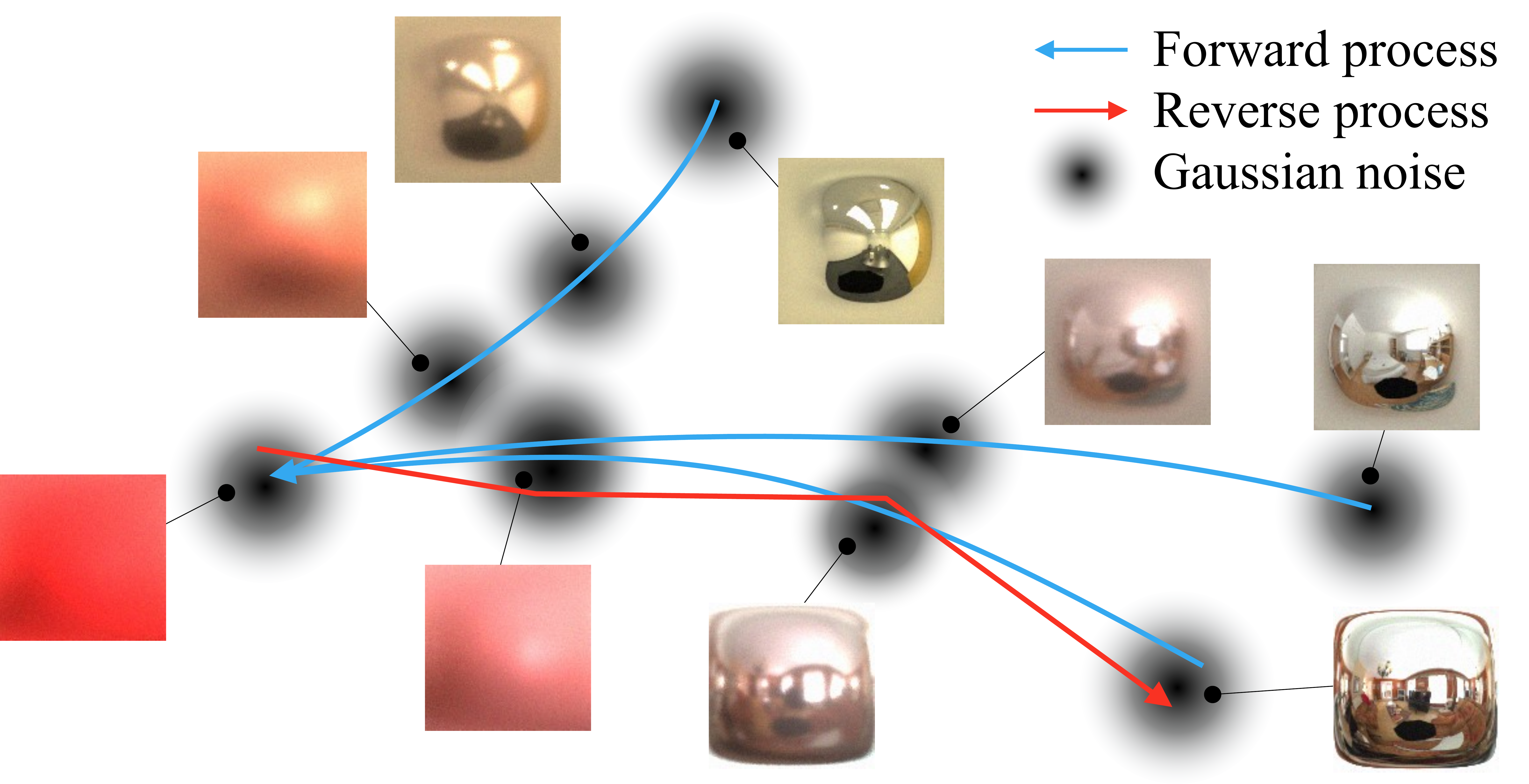}
    \caption{Additive Gaussian observation noise makes each image formation process stochastic, and reflectance attenuates high-frequencies of the illumination whose residues are gradually washed out by the noise. The reverse process is thus necessarily stochastic, effectively generating the lost higher frequency components of the illumination by sampling along learned scores.}
    \label{fig:processes_illustration}
\end{figure}

The illumination generation process by IllNet is inherently noisy due to its stochasticity and a direct estimate of the current reflectance parameters $\Psi^{(k)}$ from $L_r^{(k)}$ would be noisy especially as the step size gradually becomes smaller. 
Instead, by always estimating the observed reflectance and analytically computing the current reflectance, we obtain robust estimates of the current reflectance that are guaranteed to converge. At the same time, this lets RefNet iteratively refine its estimate of the observed object reflectance, \ie, RefNet learns to optimize the object reflectance through its iteration along with IllNet.


We can describe the radiometric formulation of an observed reflectance map $L_r^{(K)}$ from the surrounding illumination $L_r^{(0)}$ and the object reflectance $f_r^{(K)}$ as a forward iterative process 
\begin{equation}
    q(L_r^{(1:K)}| L_r^{(0)}, f_r^{(K)}) = \prod_{k=1}^K q(L_r^{(k)} | L_r^{(0)}, f_r^{(K)})\,.
    \label{eq:forward-process}
\end{equation}

Let us express the reverse process (DRMNet) using IllNet and RefNet parameters, $\theta$ and $\phi$, respectively, 
\begin{equation}
    p_\theta(L_r^{(0:K-1)}| L_r^{(K)}) = \prod_{k=1}^K p_{\theta, \phi}(L_r^{(k-1)} | L_r^{(k)})\,.
    \label{eq:reverse-process}
\end{equation}
The conditional $p_{\theta, \phi}$ must invert $q$ in \cref{eq:forward-process}. Although the forward transition $q$ is deterministic, it cannot be inverted analytically, not just because it is highly non-linear, but also because of the information loss. For this, $p_{\theta, \phi}$ must be a stochastic generative model. In order to make this reverse process stochastic, we must in turn make the forward process $q$ also stochastic while at the same time obeying the deterministic radiometric image formation process. 

Similar to \cite{rissanen2023generative}, we resolve this by adding stochastic perturbations to the deterministic forward process (\ie, the Gaussian observation noise)
\begin{equation}
    q(L_r^{(k)} | L_r^{(0)}, f_r^{(K)}) = \mathcal{N}(L_r^{(k)} | L_r(\mathbf n; L_i, f_r^{(k)}), \sigma^2 \mathbf{I})\,,
    \label{eq:forward-step}
\end{equation}
where $f_r{(k)}$ is the reflectance whose parameter $\Psi^{(k)}$ is computed from the observed reflectance estimate $\Psi^{(K)}$ by \cref{eq:refparam-def}, and $\sigma$ is the standard deviation of the added noise. 

The stochastic additive observation noise washes out the high-frequency components of the illumination, the bulk of which is already attenuated by the reflectance. \Cref{fig:processes_illustration} depicts this forward radiometric object appearance process $q$ for different combinations of illumination $L_i$ and reflectance $f_r^{(K)}$. As the reflectance deviates from perfect mirror reflection, the reflected illumination, \ie, the reflectance map, looses high-frequency details and color and all combinations become indistinguishable.
\begin{figure}[t]
    \centering
    \includegraphics[width=\linewidth]{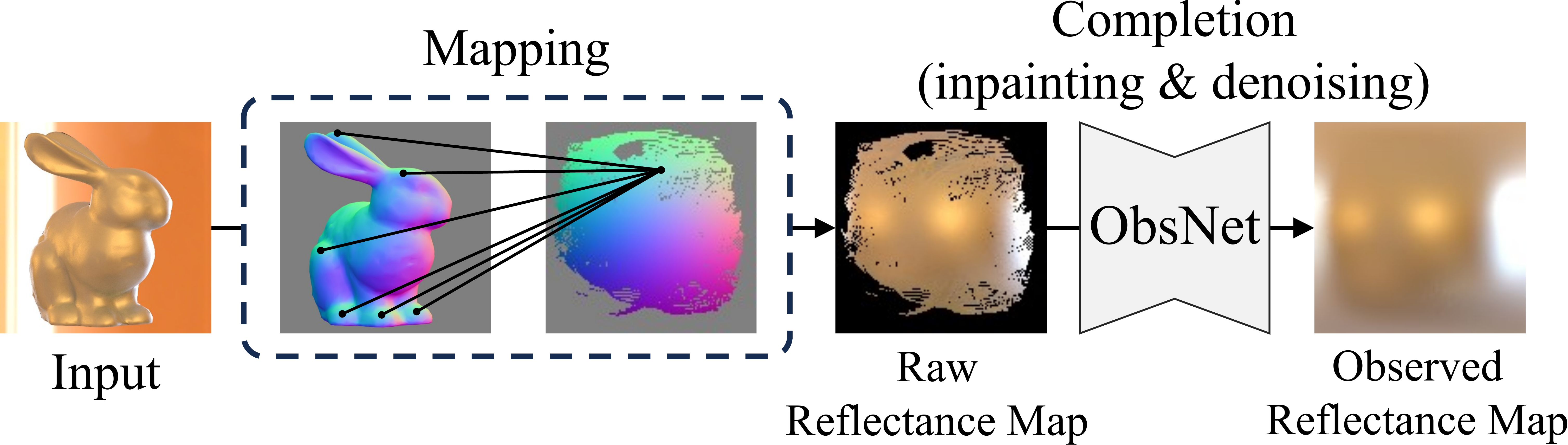}
\caption{We convert the single input image into an observation reflectance map by ``completing'' the sparsely mapped reflectance map with another diffusion model, ObsNet.} 
    \label{fig:img2refmap}
\end{figure}

The conditional $p_{\theta, \phi}$ models the reverse process of this stochastic forward radiometric process. Starting from the observed reflectance map, by iteratively reverse transitioning with the conditional $p_{\theta, \phi}$, we can generate one illumination that gives rise to the observed reflectance map for that object reflectance among the infinite possibilities. We model this reverse transition $p_{\theta, \phi}$ with a Gaussian and explicitly condition it on the observed reflectance map (\ie, object appearance) $L_r^{(K)}$
\begin{equation}
    \footnotesize{
     p_{\theta, \phi}(L_r^{(k-1)} | L_r^{(k)}, L_r^{(K)}, f_r^{(k)})
 = \mathcal{N}(L_r^{(k-1)} | \mu_{\theta, \phi}(L_r^{(k)}, L_r^{(K)}), \delta^2 \mathbf{I})\,.}
    \label{eq:reverse-step}
\end{equation}
We learn the expectation $\mu_{\theta, \phi}$ of this Gaussian 
\begin{equation*}
    \mu_\theta(L_r^{(k)}, L_r^{(K)}, \Psi_\phi^{(k)}(L_r^{(k)}, L_r^{(K)}, K - k))\,,
    \label{eq:drmnet-model}
\end{equation*}
and set the variance to a constant $\delta$.
We further reparameterize 
$\mu_\theta(L_r^{(k)}, L_r^{(K)}, \Psi_\phi^{(k)}) = L_r^{(k)} + \varepsilon_\theta(L_r^{(k)}, L_r^{(K)}, \Psi_\phi^{(k)})$
to train IllNet from the residuals.

\begin{figure*} 
    \begin{center} 
        \includegraphics[width=\linewidth]{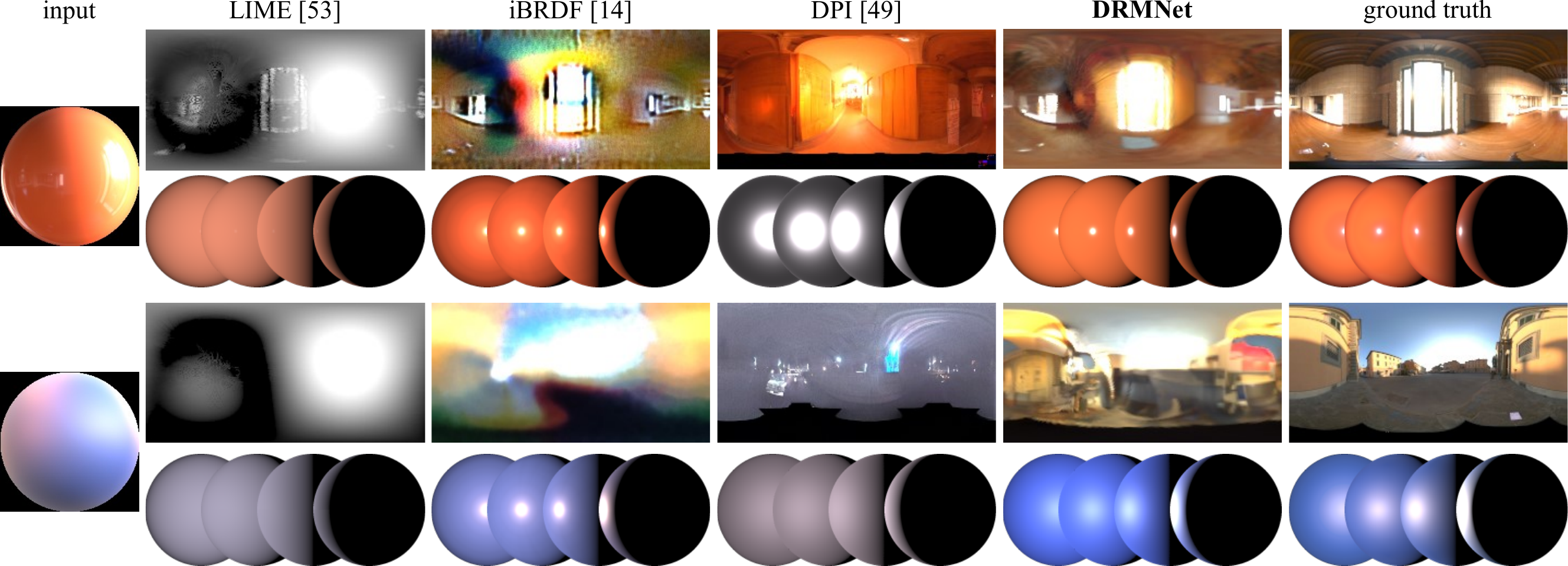} 
        \caption{Qualitative results on iBRDF synthetic dataset. For each input, the top row is the illumination estimate shown as a spherical panorama and the bottom row is the reflectance estimate rendered as a sphere under a point source.}
                \label{fig:ibrdf-synth} 
    \end{center}
    \vspace{-16pt}
\end{figure*}

We maximize the evidence lower bound (ELBO) of the marginalized reverse transition $p_{\theta, \phi}(L_r^{(0)} | L_r^{(K)})$ by minimizing the upper bound on the negative log likelihood. In the supplementary material, we derive this step by step, which leads to a simplified objective 
\begin{equation}
    \mathcal{L}_i = \mathbb{E}_{L_i, f_r, k}|\mu_{\theta,\phi}(\mathbf{n}; L_r^{(k)}, L_r^{(K)}) - L_r(\mathbf{n};L_i, f_r^{(k)})|^2_2\,.
    \label{eq:refmaploss}
    \vspace{-4pt}
\end{equation}
When training DRMNet, the illumination $L_i$, reflectance $f_r^{(K)}$, and step $k$ are uniformly sampled from a set of environment maps, reflectance parameter values, and $[1, K]$, respectively. As the reflectance map corresponding to perfect mirror reflection $L_r^{(0)}$ is equivalent to the incident illumination $L_i$, we can use $L_i$ as the expectation instead of $L_r^{(0)}$. 

In addition to the loss on the reflectance map \cref{eq:refmaploss}, we impose a reflectance loss on the reflectance parameters estimated by RefNet $\Psi_\phi^{(K)}$ using the reflectance parameters corresponding to the forward step and process, \cref{eq:forward-step} $f_r^{(k)}$ and \cref{eq:forward-process} $f_r^{(K)}$, respectively, 
\begin{equation}
    \mathcal{L}_r = \mathbb{E}_{L_i, f_r, k}\left[|\Psi_\phi^{(k)} - \Psi^{(k)}|^2_2 + |\Psi_\phi^{(K)} - \Psi^{(K)}|^2_2\right]\,.
    \label{eq:refloss}
\end{equation}


The final loss is 
    $\mathcal{L}=\lambda_i \mathcal{L}_i + \lambda_r \mathcal{L}_r$,
where $\lambda_i$ and $\lambda_r$ are hyperparameters weighting the ELBO and reflectance losses. IllNet is implemented with a U-Net and RefNet with a simple MLP. Please see the supplementary material for detailed architectures of IllNet and RefNet.


\section{Observed Reflectance Map}
\label{sec:obsref}

\begin{figure*} 
    \begin{center} 
        \includegraphics[width=\linewidth]{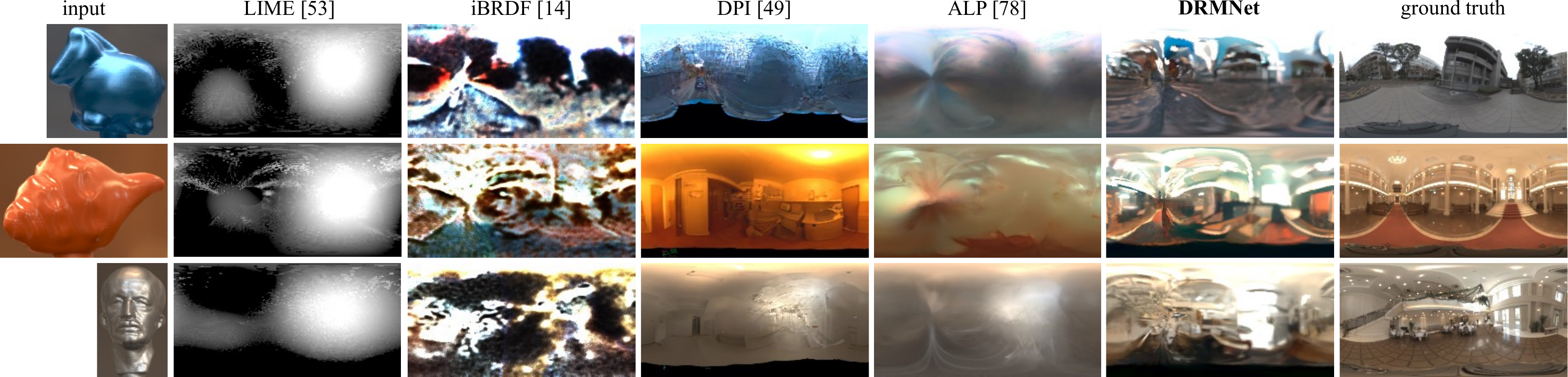} 
        \caption{Illumination estimates of the nLMVS-Real dataset for different objects taken in complex environments. DRMNet successfully recovers accurate and plausible detailed illumination from the frequency-attenuated object appearance. Note that ALP knows the reflectance.}
                \label{fig:nlmvs-real_illum} 
    \end{center}
    \vspace{-16pt}
\end{figure*}
How do we get the first input to DRMNet from the single input image? Given the single image of an object whose geometry as surface normals are known, we can compute its reflectance map by mapping each pixel onto a Gaussian hemisphere, \ie a hemisphere parameterized by the surface normals with its north pole pointing towards the viewpoint. Each pixel $p$ gives a pair of surface radiance $L_p$ and surface normal $\mathbf{n}_p$ related by \cref{eq:refmap}
\begin{equation}
    L_p = L_r(\mathbf{x}_p, \mathbf{R}_{\mathbf{n_p}}w_{o,p})\,,
    \label{eq:perspective-pixel}
\end{equation}
where $\mathbf{x}_p$ is the surface point of pixel $p$. We assume orthographic projection and use the camera coordinate frame so that $w_{o,p}=[0,0,-1]$ and a homogeneous surface $L_p = L_r(\mathbf{n_p})$. The reflectance map $L_r(\mathbf{n_p})$ can thus be computed by extracting the surface normal of each image pixel.

As each surface normal of the reflectance map can be observed at multiple image pixels, the mapping from the input image to the reflectance map is not injective. 
Global light transport manifests in self-shadows and interreflection causing darker and brighter radiance, respectively, for the same surface normal at other surface points that are directly lit. 
We minimize global light transport effects in our reflectance map mapping by taking the median pixel intensities of the multiple surface points with the same surface normal 
\begin{equation}
    \boldsymbol r(\boldsymbol n) = \mathrm{median}\{L_p | \theta_p := \cos^{-1}(\boldsymbol n_p\cdot \boldsymbol n) < \epsilon\}\,,
    \label{eq:mapping}
\end{equation}
where $\epsilon$ is the angular threshold to determine the set of pixels corresponding to a given surface normal.

\begin{table}[t]
    \centering
    \small 
    \setlength\tabcolsep{1.5pt} 
    \begin{tabular}{lcc}
                        & \multicolumn{1}{c}{illumination} & \multicolumn{1}{c}{reflectance} \\
                        & logRMSE\textdownarrow/PSNR\textuparrow/SSIM\textuparrow/LPIPS\textdownarrow              & logRMSE\textdownarrow  \\\hline
        LIME \cite{meka2018lime}       & 5.74~/~~~8.7~/~0.24~/~0.70      & 1.88                          \\
        iBRDF \cite{chen2022invertible}& 3.20~/~11.5~/~0.34~/~0.67      & 1.01                     \\
        DPI \cite{lyu2023dpi}          & 4.22~/~10.8~/~0.25~/~0.68      & 2.03                    \\
        DRMNet (Ours)                  & \textbf{2.62}~/~\textbf{14.4}~/~\textbf{0.59}~/~\textbf{0.59}      & \textbf{0.66}                         \\
    \end{tabular}
    \vspace{-4pt}
    \caption{Quantitative results on the iBRDF synthetic dataset \cite{chen2022invertible}. Our DRMNet acheives state-of-the-art accuracy.}
    \label{tab:ibrdf-synth}
\end{table}
This reflectance map is inevitably sparse unless the surface normals of the object cover all possible directions uniformly, \ie, it is a hemisphere facing the camera. We ``complete'' this reflectance map with another diffusion model, which we refer to as ObsNet. 
Unlike regular diffusion for image inpainting \cite{bansal2022cold}, we train the diffusion model to learn to generate the observed regions as they can be inaccurate due to global light transport effects that slipped through the median mapping. ObsNet can be trained on a large number of synthetic reflectance maps rendered for various combinations of reflectance and illumination masked by surface normal distributions of randomly selected objects.

\begin{table}[t]
    \centering
    \small 
    \setlength\tabcolsep{1.5pt} 
    \begin{tabular}{lcccccccc}
                                           & \multicolumn{1}{c}{illumination} & est. time                              \\
                                           & logRMSE\textdownarrow/PSNR\textuparrow/SSIM\textuparrow/LPIPS\textdownarrow              &  secs\textdownarrow \\ \hline
        LIME \cite{meka2018lime}           & 7.05~/~8.11~/~0.14~/~0.71             & \textbf{1.1}   \\
        iBRDF \cite{chen2022invertible}    & 2.13~/~10.6~/~0.17~/~0.70             & 5048  \\
        DPI \cite{lyu2023dpi}              & 3.57~/~12.6~/~0.27~/~\underline{0.65}              & 1143  \\
        ALP \cite{yu2023accidental}        & \textbf{0.95}~/~\textbf{15.3}~/~\textbf{0.39}~/~0.66              & 267   \\
        DRMNet (Ours)                      & \underline{1.18}~/~\underline{14.6}~/~\underline{0.33}~/~\textbf{0.60}              & \underline{14.3}  \\
    \end{tabular}
    \vspace{-4pt}
    \caption{Quantitative evaluation of the illumination estimates on the nLMVS-Real dataset \cite{yamashita2023nlmvs}. Note that ALP requires reflectance which we provide by estimation from multiview images a priori. DRMNet achieves comparable accuracy 20 times faster while jointly estimating the reflectance from a single image.}
    \label{tab:nlmvs-real_illum}
\end{table}

\section{Experiments}
\label{sec:exp}

We evaluate the effectiveness of DRMNet quantitatively and qualitatively and compare with related existing inverse-rendering methods \cite{meka2018lime, chen2022invertible, yu2023accidental,lyu2023dpi} on synthetic and real images. In the supplementary material, we also justify our formulation of stochastic inverse rendering with probabilistic diffusion as well as the specific architecture of DRMNet through ablation studies, and analyze the stochastic behavior of DRMNet over multiple estimations on the same input. Please see supplementary material for more results.

\begin{table}[t]
    \centering
    \footnotesize
    \begin{tabular}{lccccc}
                                                      & \multicolumn{1}{c}{illumination} & \multicolumn{1}{c}{relighting}                    \\
                                                      & logRMSE\textdownarrow/DSSIM\textdownarrow                           & logRMSE\textdownarrow/DSSIM\textdownarrow   \\ \hline
        DelightNet & 0.933~/~0.365                & 1.110~/~0.186    \\
        iBRDF      & 0.864~/~0.329                & 1.027~/~\textbf{0.077}    \\
        Ours       & \textbf{0.694}~/~\textbf{0.308}                & \textbf{0.414}~/~0.107    \\
    \end{tabular}
    \vspace{-4pt}
    \caption{Quantitative results on the DeLight-Net dataset. DRMNet consistently achieves high accuracy in both illumination and reflectance estimation. See text for details.}
    \label{tab:delight-real_illum}
\end{table}

\vspace{-8pt}
\paragraph{Datasets}
We train DRMNet on a large dataset of synthetic reflectance maps rendered with Mitsuba3 \cite{Mitsuba3}. We use environment maps from the Laval Indoor Dataset \cite{gardner2017learning, bolduc2023beyond} and Poly Haven HDRIs \cite{polyhaven} for illumination. Please see the supplementary material for details about how we randomly sample these illumination and reflectance to create a large training and test dataset. We use the same dataset to train ObsNet. All data and code can be found in the project page.

For thorough comparative studies, we use the iBRDF synthetic dataset \cite{chen2022invertible}, the nLVMS real dataset \cite{yamashita2023nlmvs}, and the DeLight-Net real dataset \cite{georgoulis2018reflectance}. The iBRDF dataset consists of images of spheres rendered with HDR environment maps \cite{debevec2008rendering} and measured BRDFs \cite{Matusik2003datadriven}. This dataset allows us to quantitatively evaluate the accuracy of reflectance estimates, which is not possible with real images. 
We use a large enough subset of this dataset to achieve thorough quantitative comparison in realistic time. 
The nLMVS dataset consists of images captured in six different illumination environments for 20 different objects of 5 different shapes and 4 different reflectances, all with ground truth geometry and HDR environment maps. We test on one view for each environment. The DeLight-Net dataset consists of images capturing real spheres and is used to compare with the method by Georgoulis \etal \cite{georgoulis2018reflectance} as their code is not available (link broken).

\begin{figure}[t]
    \centering
    \includegraphics[width=\linewidth]{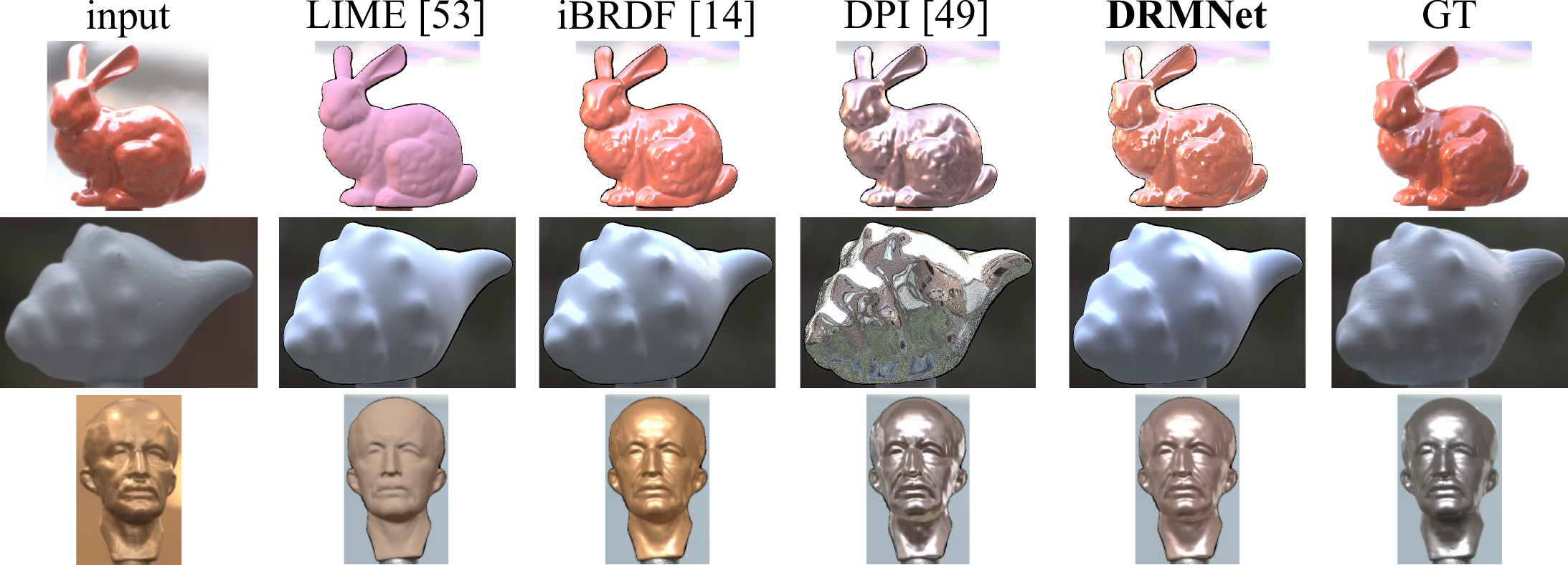}
    \caption{Relighting results using reflectance estimates on the nLMVS-Real dataset. DRMNet results match ground truth well, demonstrating the accuracy of its recovered illumination. ALP cannot be applied as it requires reflectance pre-acquisition.}
    \label{fig:nlmvs-real_relight}
\end{figure}

\begin{figure}[t]
    \centering
    \includegraphics[width=\linewidth]{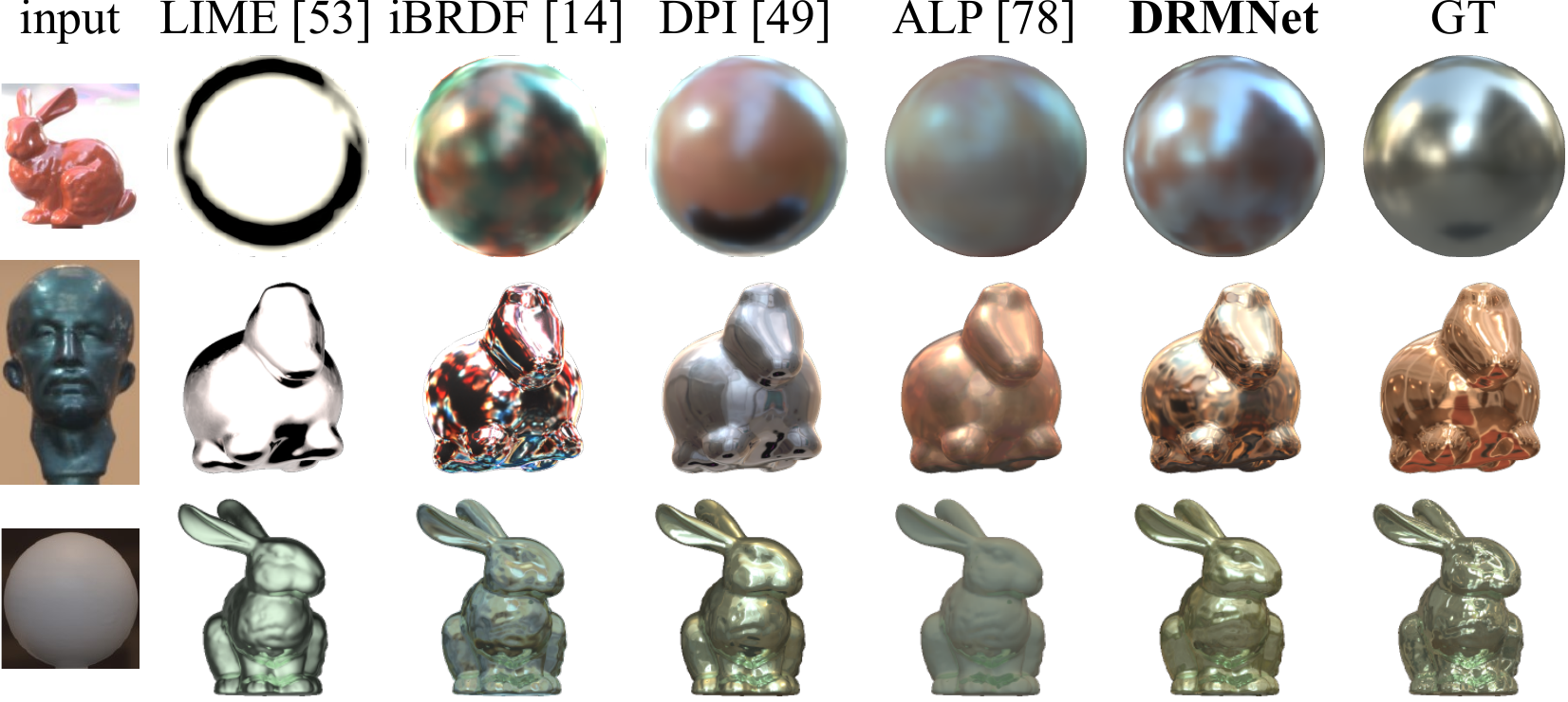}
    \caption{Object replacement results on the nLMVS-Real dataset. Accurate illumination estimates of DRMNet enable rendering of detailed object appearance for objects with arbitrary reflectance.}
    \label{fig:nlmvs-real_objreplace}
    \vspace{-4pt}
\end{figure}
\vspace{-8pt}
\paragraph{Metrics}
We use several metrics to quantitatively evaluate the accuracy of the illumination and reflectance estimates. For the reflectance, we use the log-scale RMSE in the non-parametric MERL BRDF representation as in iBRDF \cite{chen2022invertible}. The non-parametric representation enables accuracy comparison regardless of the BRDF model. 
Scale-invariance absorbs the ambiguity in absolute radiance due to exposure and overall illumination brightness. 
We also evaluate reflectance estimates through relighting results with log-scale RMSE and SSIM \cite{wang2004image}. For quantitative evaluation of the illumination estimates, we compute scale-invariant log-scale RMSE, PSNR, SSIM, and LPIPS \cite{zhang2018unreasonable} at $128\times 256$ resolution. For the PSNR, SSIM, and LPIPS, we compute them on LDR images after tone-mapping HDR illumination.


\vspace{-8pt}
\paragraph{iBRDF synthetic dataset}
\Cref{tab:ibrdf-synth} and \cref{fig:ibrdf-synth} show quantitative and qualitative evaluation of the accuracy of illumination and reflectance estimates using the iBRDF synthetic dataset. DRMNet achieves higher accuracy and recovers more natural illumination. As ALP \cite{yu2023accidental} requires estimation of the shape and reflectance parameters at many viewpoints, we do not compare with it here. \Cref{fig:ibrdf-synth} shows that the concurrent work by Lyu \etal \cite{lyu2023dpi} results in illumination estimates that significantly deviate from the ground truth, as it is a naive noise seeded diffusion model used as an external prior in a classic Bayesian inverse-rendering formulation.

\vspace{-8pt}
\paragraph{nLMVS real dataset}
We evaluate the estimation accuracy on real images using the nLMVS-Real dataset \cite{yamashita2023nlmvs}.
\Cref{tab:nlmvs-real_illum} shows quantitative results. Note that we have also included ALP \cite{yu2023accidental} for comparison, but the method requires known reflectance, pre-acquired from multiple images under known lighting. 
Our method achieves accuracy comparable to this known-reflectance method, but in an order faster computation as DRMNet does not require complex non-linear optimization. 

As ground truth reflectance is unknown, we evaluate the accuracy of the reflectance estimate through relighting. \Cref{fig:nlmvs-real_relight} shows relighting results under a different illumination using the estimated reflectance and rotated geometry to align with that in the ground truth image. The results show strong consistency in object appearance, suggesting high accuracy of reflectance estimates. 
\Cref{fig:nlmvs-real_objreplace} shows renderings of different objects under the estimated illumination. Since our method explicitly recovers the high frequency spectrum of the illumination, even objects with higher frequency than the one used to recover the illumination can be relit with natural appearance. Please see supplementary material for more results in higher resolution.
\Cref{tab:nlmvs-real_app} shows quantitative results corresponding to \cref{fig:nlmvs-real_relight,fig:nlmvs-real_objreplace}. DRMNet achieves highest accuracy in both.

\begin{table}[t]
    \centering
    \setlength{\tabcolsep}{2.7pt}
    \footnotesize
    \begin{tabular}{lcc}
        & \multicolumn{2}{c}{logRMSE\textdownarrow/PSNR\textuparrow/SSIM\textuparrow/LPIPS\textdownarrow} \\
        & relighting & object replacement \\\hline
        LIME \cite{meka2018lime}           & 1.73~/~18.6~/~\underline{0.81}~/~0.40 & 1.67~/~14.0~/~0.73~/~0.25          \\ 
        iBRDF \cite{chen2022invertible}     & \underline{1.69}~/~\underline{19.9}~/~\textbf{0.83}~/~\underline{0.35} & 0.58~/~19.2~/~0.80~/~0.22  
            \\ 
        DPI \cite{lyu2023dpi}               & 1.75~/~17.7~/~0.73~/~0.44 & 0.49~/~21.0~/~\underline{0.83}~/~0.21 \\ 
        ALP \cite{yu2023accidental};         & (pre-acquired reflectance) & \underline{0.39}~/~\underline{22.2}~/~\textbf{0.86}~/~\underline{0.19} \\ 
        \textbf{DRMNet}                        & \textbf{1.68}~/~\textbf{20.6}~/~\textbf{0.83}~/~\textbf{0.34}  & \textbf{0.35}~/~\textbf{23.3}~/~\textbf{0.86}~/~\textbf{0.17} \\
    \end{tabular}
    \vspace{-4pt}
    \caption{Quantitative evaluation of relighting with rotated geometry and object replacement on the nLMVS-Real dataset \cite{yamashita2023nlmvs}. Our method achieves highest accuracy in both.}
    \label{tab:nlmvs-real_app}
\end{table}


\vspace{-8pt}
\paragraph{DeLight-Net real dataset}
\Cref{tab:delight-real_illum} shows quantitative results on the DeLightNet dataset. ``Illumination'' is the estimated illumination reflected by a mirror sphere and Georgoulis \etal \cite{georgoulis2018reflectance} directly regress this from the reflectance map. Note that DeLightNet does not estimate the reflectance. To directly compare with their method, we evaluate using log-scale RMSE and DSSIM \cite{dssim}. Our DRMNet achieves higher accuracy over the two existing methods in both illumination and reflectance estimates except for one metric. We plan to incorporate a learned BRDF model, \eg, iBRDF \cite{chen2022invertible}, in future extensions.


\section{Conclusion}

We introduced DRMNet, a principled stochastic inverse rendering method that estimates the illumination and reflectance from a single image of an object of known geometry taken under complex natural illumination. By formulating the radiometric disentanglement on the reflectance map and as a recursive inversion of stochastic diffusion but with an underlying deterministic rendering equation, we showed that DRMNet achieves state-of-the-art accuracy and full detailed recovery of surrounding illumination, enabling relighting and replacement of objects with arbitrary reflectance properties. By breaking the fundamental limitations of inverse rendering with a seamlessly integrated generative model, DRMNet opens new possibilities of single-image radiometric understanding.

\vspace{-8pt}
\paragraph{Acknowledgements}
This work was in part supported by
JSPS 
20H05951, 
21H04893, 
JST JPMJCR20G7 
and JPMJAP2305, 
and RIKEN GRP.

{
\small
\bibliographystyle{ieeenat_fullname}
\bibliography{cvpr24_yenyo_arxiv}
}

\setcounter{section}{0}
\input{cvpr24_yenyo_arxiv_supp_cr}

\end{document}

%% file: preamble.tex
\usepackage[dvipsnames]{xcolor}


%% file: cvpr24_yenyo_arxiv_supp_cr.tex
\clearpage
\setcounter{page}{1}
\maketitlesupplementary

\renewcommand{\thesection}{\Alph{section}}

References in this supplementary material refer to the citation numbers in the main text for those citations already made in the paper. For new references, we continue the numbering from the main text. 

\section{The Reflectance Model}

We describe the full reflectance model we use in DRMNet which is based on the Disney principled BRDF model \cite{Burley12}. The model is composed of diffuse reflection $f_\mathrm{diff}$, retro reflection $f_\mathrm{retro}$, and specular reflection $f_\mathrm{spec}$.
The diffuse reflection $f_\mathrm{diff}$ depends on the angle of incidence $\theta_\mathrm{i}$ and that of outgoing direction $\theta_\mathrm{o}$
\begin{equation}
    f_\mathrm{diff} = (1 - \frac{F_\mathrm{i}}{2})(1 - \frac{F_\mathrm{o}}{2}),
\end{equation}
where $F_{\mathrm{i}} = (1 - \cos\theta_{\mathrm{i}})^5$ and $F_{\mathrm{o}} = (1 - \cos\theta_{\mathrm{o}})^5$.
The retro reflection $f_\mathrm{retro}$ is defined by the metallic parameter $\gamma$ 
\begin{equation}
    f_\mathrm{retro} = R_R (F_\mathrm{i} +  F_\mathrm{o} + F_\mathrm{i}F_\mathrm{o}(R_R - 1)),
\end{equation}
where $R_R = 2\gamma\cos^2\theta_\mathrm{d}$ and $\theta_\mathrm{d}$ is the angle between the half vector $\boldsymbol{h}$ and the incident direction.
The specular reflection $f_\mathrm{spec}$ is based on the microfacet model with the GGX distribution
\begin{equation}
    f_\mathrm{spec} = \frac{FDG}{4\cos\theta_\mathrm{i}\cos \theta_\mathrm{o}}\,,
\end{equation}
where $D$ is a microfacet distribution function defined by the roughness parameter $\alpha$ 
\begin{equation}
    D = \frac{\alpha^4}{\pi((\boldsymbol{h}\cdot \boldsymbol{n})^2(\alpha^4 - 1) + 1)^2}\,,
\end{equation}
$\boldsymbol{n}$ is the normal direction of the surface in the local frame, and $G$ is the shadowing-masking function 
\begin{equation}
    G_1(\boldsymbol\omega) = \dfrac{2}{1 + \sqrt{1 + \alpha^4 (1 - \boldsymbol\omega\cdot \boldsymbol{n})^2 / (\boldsymbol\omega \cdot \boldsymbol{n})^2}}\,,
\end{equation}
and $G = G_1(\boldsymbol\omega'_\mathrm{i})G_1(\boldsymbol\omega'_\mathrm{o})$.
$F$ is the Fresnel term which is a combination of
\begin{equation}
    F_{\mathrm{dielectric}} = \frac{\left(\dfrac{\cos \theta_\mathrm{o} - \eta \cos \theta_\mathrm{t}}{\cos \theta_\mathrm{o} + \eta \cos \theta_\mathrm{t}}\right)^2 + \left(\dfrac{\cos \theta_\mathrm{t} - \eta \cos \theta_\mathrm{o}}{\cos \theta_\mathrm{t} + \eta \cos \theta_\mathrm{o}}\right)^2}{2}\,,
\end{equation}
where $\eta = \dfrac{2}{1 - \sqrt{0.08\rho_\mathrm{s}}} - 1$, and
\begin{equation}
    F_{\mathrm{Schlick}} = \boldsymbol{\rho}_\mathrm{d} + (1 - \boldsymbol{\rho}_\mathrm{d})(1-\cos \theta_\mathrm{d})^5\,,
\end{equation}
and
\begin{equation}
    F=(1 - \gamma)F_{\mathrm{dielectric}} + \gamma F_{\mathrm{Schlick}}.
\end{equation}

\section{Objective Function Derivation}
Let us derive the objective function of the reflectance map \cref{eq:refmaploss}
step by step. The objective of the reverse process is to maximize the marginalized likelihood of the reverse process $p_{\theta, \phi}(L_r^{(0)} | L_r^{(K)})$. Instead of directly maximizing this likelihood, we minimize the negative log likelihood
\begin{align}
    &- \log p_{\theta, \phi}(L_r^{(0)}|L_r^{(K)})\nonumber\\
    &\leq \mathbb E_{q}\left[ -\log \frac {p_{\theta, \phi}(L_r^{(0:K-1)}|L_r^{(K)})} {q(L_r^{(1:K-1)} | L_r^{(0)}, L_r^{(K)}, f_r^{(K)})}\right]=:\mathcal{L}_p\,,
    \label{eq:elbo}
\end{align}
Here, $\mathcal{L}_p$ is
\begin{align}
    \mathcal{L}_p &=\mathbb E_{q}[ -\sum_{k=2}^K \log \frac {p_{\theta, \phi}(L_r^{(k-1)} | L_r^{(k)}, L_r^{(K)})} {q(L_r^{(k-1)} | L_r^{(0)}, L_r^{(K)}, f_\mathrm{r}^{(K)})}\nonumber\\
    &-\log p_{\theta, \phi}(L_r^{(0)}|L_r^{(1)}, L_r^{(K)})] \nonumber\\ 
    &= \sum_{k=2}^K \mathbb E_{q_{/\{k-1\}}} D_{KL}[ q(L_r^{(k-1)} | L_r^{(0)}, L_r^{(K)}, f_\mathrm{r}^{(K)})\nonumber\\
    &|| p_{\theta, \phi}(L_r^{(k-1)} | L_r^{(k)}, L_r^{(K)})] \nonumber\\
    &- \mathbb E_{q} \log p_{\theta, \phi}(L_r^{(0)}|L_r^{(1)}, L_r^{(K)}) \nonumber\\
    &= \sum_{k=2}^K \mathbb E_{q_{k}} D_{KL}[ q(L_r^{(k-1)} | L_r^{(0)}, L_r^{(K)}, f_\mathrm{r}^{(K)})\nonumber\\
    &|| p_{\theta, \phi}(L_r^{(k-1)} | L_r^{(k)}, L_r^{(K)})] \nonumber\\
    &- \mathbb E_{q_1} \log p_{\theta, \phi}(L_r^{(0)}|L_r^{(1)}, L_r^{(K)})\,,
    \label{eq:detailed-lp}
\end{align}
where 
\begin{align}
    \mathbb E_{q} &= \mathbb E_{q(L_r^{(1:K-1)}|L_r^{(0)}, L_r^{(K)}, f_\mathrm{r}^{(K)})}\\
    \mathbb E_{q_{/\{k-1\}}} &= \mathbb E_{q(L_r^{\{1:K-1\}/\{ k-1\}} | L_r^{(0)}, L_r^{(K)}, f_\mathrm{r}^{(K)})}\\
    \mathbb E_{q_{k}} &= \mathbb E_{q(L_r^{(k)}|L_r^{(0)}, L_r^{(K)}, f_\mathrm{r}^{(K)})}\\
    \mathbb E_{q_1} &= \mathbb E_{q(L_r^{(1)}|L_r^{(0)}, L_r^{(K)}, f_\mathrm{r}^{(K)})}
\end{align}
By modeling the forward process \cref{eq:forward-step}
and its reverse process \cref{eq:reverse-step}
the first term of \cref{eq:detailed-lp} becomes
\begin{align}
    &D_{KL}\mathcal{N} (L_r^{(k)}|R(\mathbb \ell, f_\mathrm r^{(k)}), \sigma^2 \boldsymbol I) \nonumber\\
    &||\mathcal N(L_r^{(k-1)} | \boldsymbol \mu_{\theta, \phi}(L_r^{(k)}, L_r^{(K)}, \boldsymbol z^{(k)}), \delta^2 \mathbf I)\nonumber\\
    &= \frac{1}{2}[2N \log \frac{\delta}{\sigma} - N +\frac{1}{\sigma^2} \nonumber\\
    &||R(\mathbb \ell, f_\mathrm r^{(k)}) - \mu_{\theta, \phi}(L_r^{(k)}, L_r^{(K)}, \boldsymbol z^{(k)})||_2^2 + \frac{\sigma^2}{\delta^2}N]\,,
\end{align}
and the second term becomes 
\begin{align}
    &\frac{1}{2\delta^2}\mathbb E_{q(L_r^{(1)}|L_r^{(0)}, L_r^{(K)}, f_\mathrm r)} \left[||\mu_{\theta, \phi}(L_r^{(1)}, L_r^{(K)}, \boldsymbol z^{(1)}) - L_r^{(0)}||_2^2\right] \nonumber\\
    &+ N\log(\delta\sqrt{2\pi})\,.
\end{align}
By focusing on those terms related to the model parameters $\theta, \phi$, we obtain the simplified objective, \cref{eq:refmaploss}.

\begin{figure}[t]
    \centering
    \includegraphics[width=\linewidth]{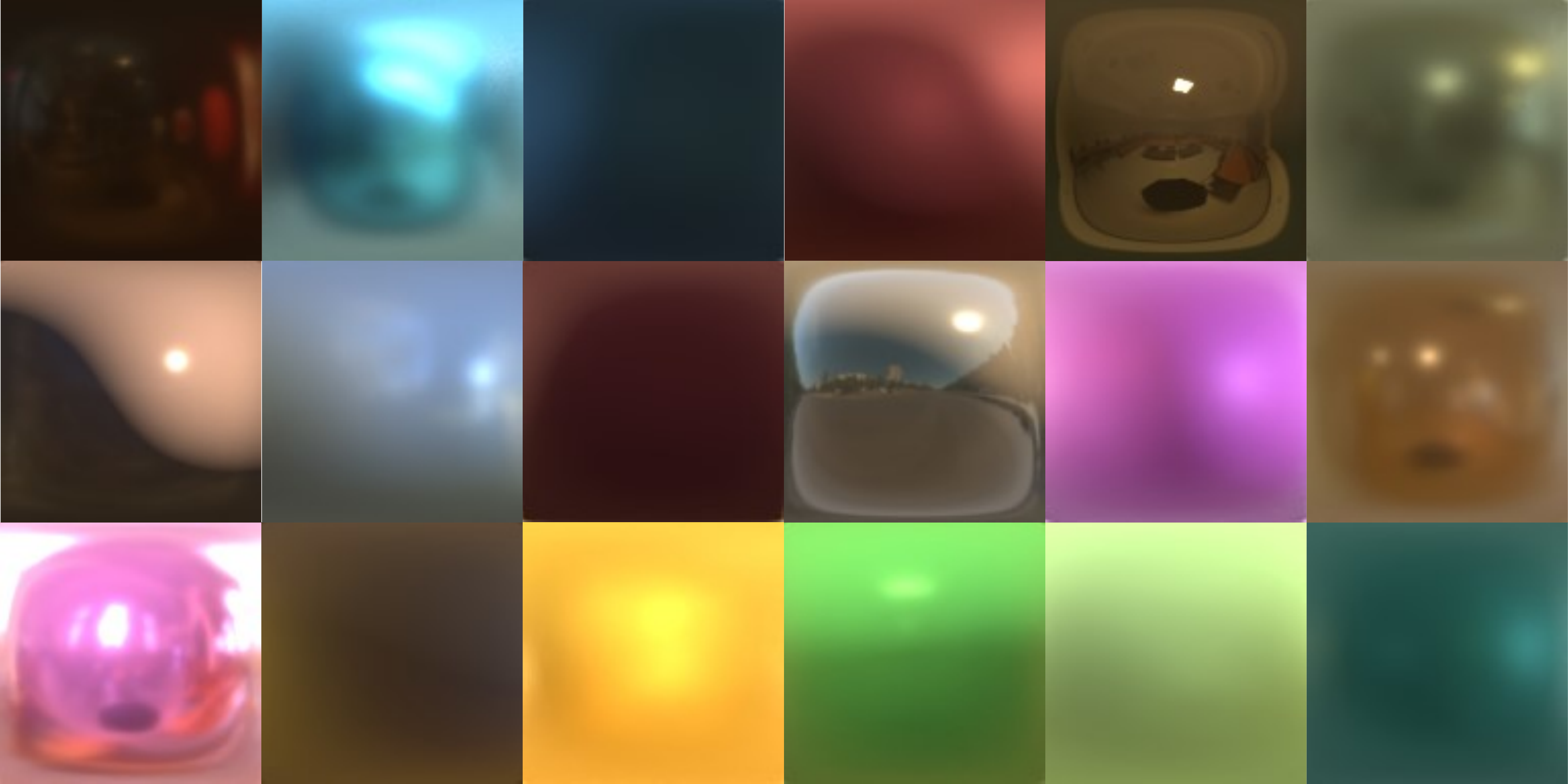}
    \caption{Samples from the synthetic reflectance map dataset we create to train DRMNet. The dataset consists of various combinations of HDR environment maps and reflectance.}
    \label{fig:training-samples}
\end{figure}

\section{Network Architectures}

As in regular probabilistic diffusion models \cite{Ho20, rombach2022highresolution}, we use a U-Net with skip connections as the network architecture for IllNet which we denote as $\varepsilon_\theta$ in $\mu_\theta(L_r^{(k)}, L_r^{(K)}, \Psi_\phi^{(k)}) = L_r^{(k)} + \varepsilon_\theta(L_r^{(k)}, L_r^{(K)}, \Psi_\phi^{(k)})$.
The input to IllNet is the concatenated observed reflectance map and current reflectance map, $L^{(K)}$ and $L^{(k)}$, respectively.
Each layer of the encoder and the decoder contains two residual blocks and consists of 1 to 6 times of 128 channels in increment of 1 from the highest resolution layer. 
Each residual block additively embeds the current reflectance parameter $\Psi^{(k)}$ to the feature map. 
In contrast to a regular probabilistic diffusion model which uses sinusoidal positional encoding of the time step together with an MLP to compute the embedding vector, we directly use an MLP to compute the embedding vector from $\Psi^{(k)}$.
The input observed reflectance map is $128\times 128$ in resolution. At low-resolution layers of $16\times 16$, $8\times 8$, and $4\times 4$, we apply self-attention to the feature map after the residual block. 

For RefNet, we use an encoder of a U-Net with an MLP. As the same as IllNet, the input to RefNet is the concatenated observed reflectance map $L^{(K)}$ and the current reflectance map $L^{(k)}$, and the output is a 6-dimensional vector $\Psi^{(K)}$ of the observed object's reflectance parameter.
Each layer of the encoder consists of two residual blocks and has 1, 1, 2, 3, and 4 times of 128 channels from the highest-resolution. 
The encoder uses the traditional sinusoidal positional encoding and an MLP to additively embed the time steps taken so far $(K-k)$ into each residual block. 
At the lower-resolution layers of $16\times 16$ and $8\times 8$, we use self-attention. 

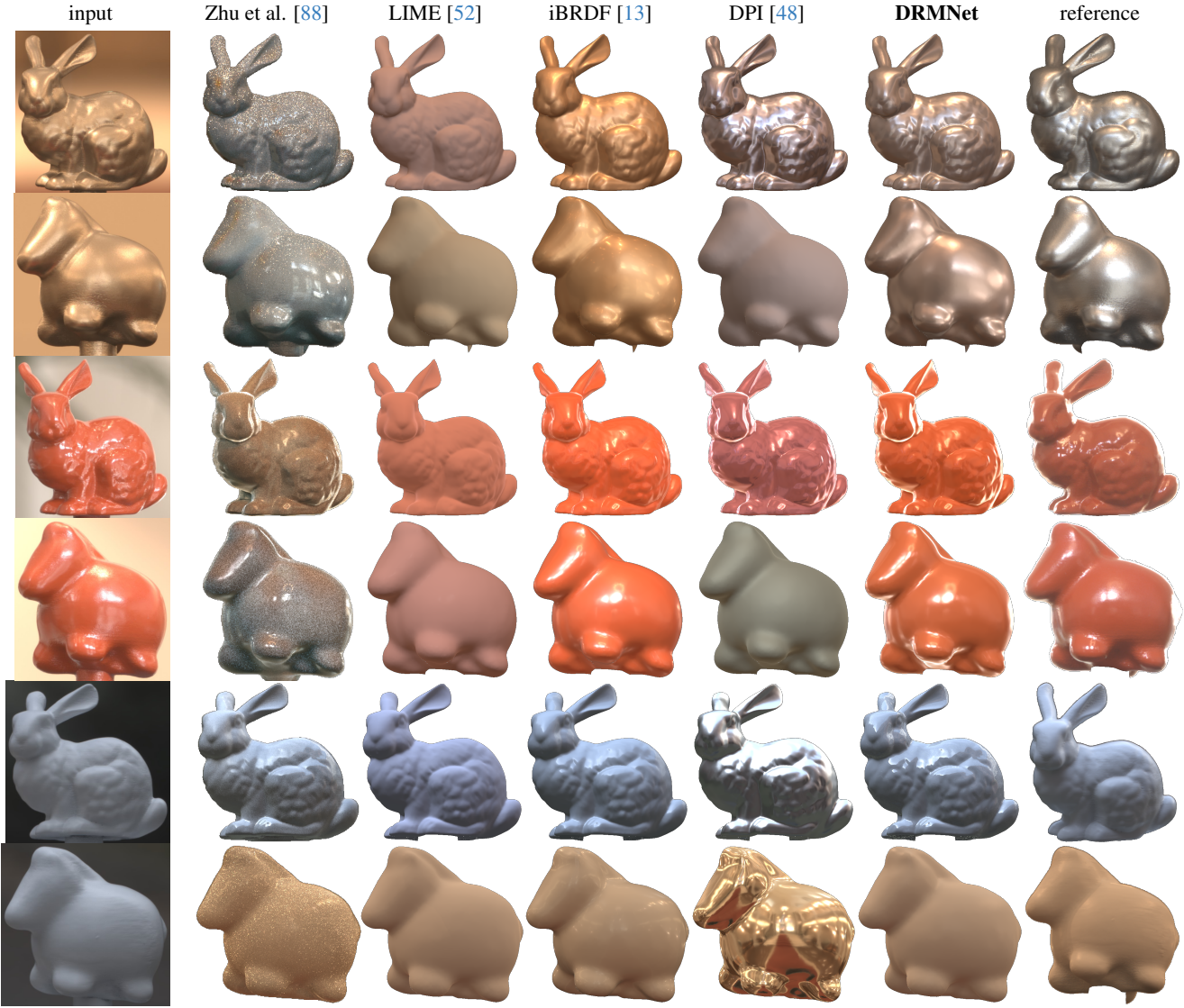
\begin{figure*}[t]
    \centering
    \def\svgwidth{\linewidth}
    \input{figs/nlmvs-real_relighting_additional_sep}
    \vspace{-14pt}
    \caption{Additional relighting results for the nLMVS-Real dataset \cite{yamashita2023nlmvs}. DRMNet achieves higher qualitative accuracy suggesting its superior accuracy of reflectance estimates.}
    \label{fig:additional-nlmvs-relight}
\end{figure*}
\begin{figure*}[t]
    \centering
    \def\svgwidth{\linewidth}
    \input{figs/nlmvs-real_objreplace_additional_sep}
    \vspace{-14pt}
    \caption{Additional object replacement results for the nLMVS-Real dataset \cite{yamashita2023nlmvs}. DRMNet results in qualitatively higher accuracy suggesting its superior accuracy of illumination estimates.}
    \label{fig:additional-nlmvs-objreplace}
\end{figure*}
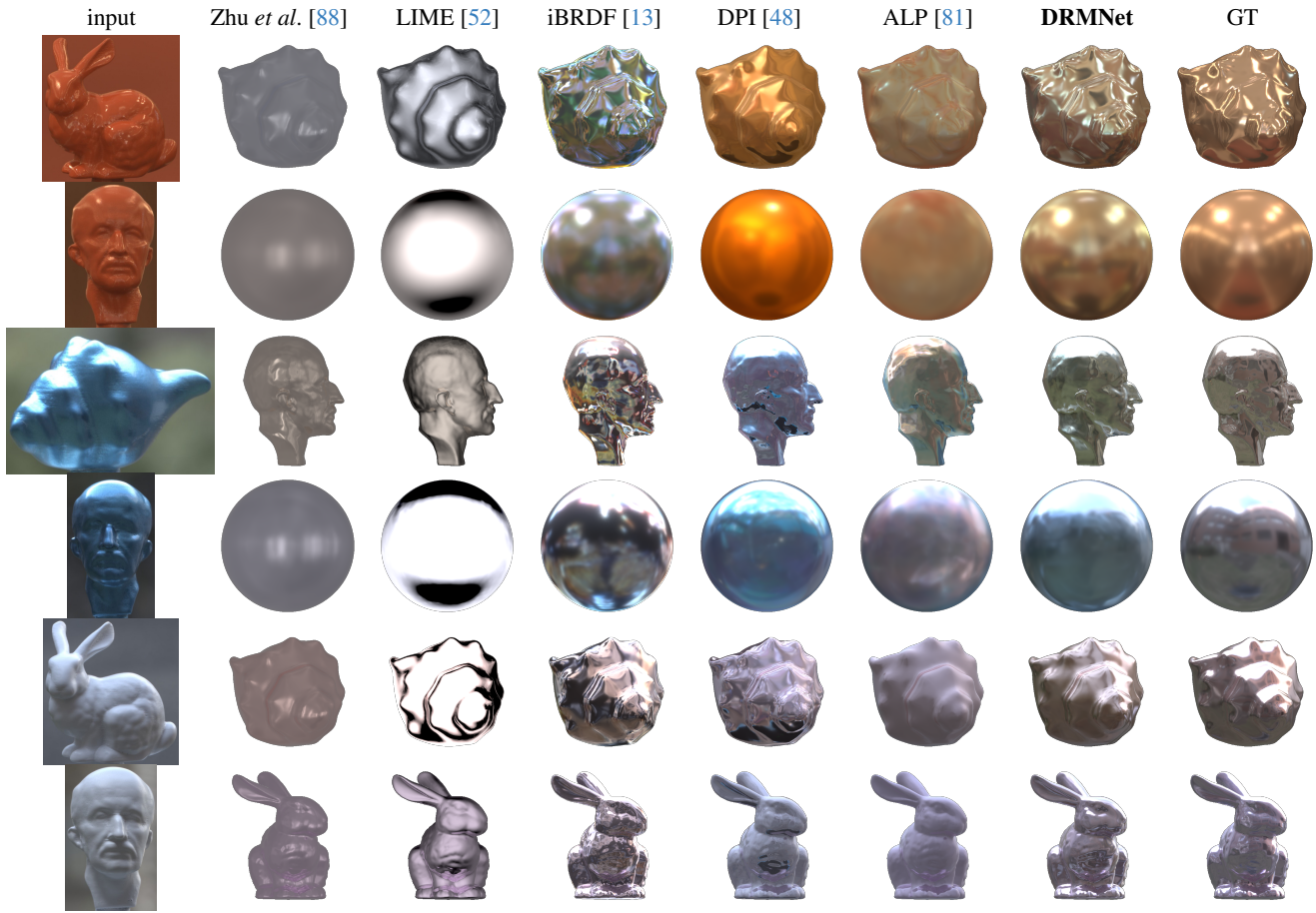
For ObsNet, we use a U-Net with 1, 2, 3, 4, and 5 times of 128 channels at each layer and self-attention at the $16\times 16$ and $8\times 8$ low-resolution layers. Similar to existing diffusion inpainting methods \cite{saharia2022palette}, the network learns to inpaint through inverse diffusion by conditioning on the sparse raw observed reflectance map with missing regions filled with noise through concatenation to the input noise at each step.
Inspired by latent diffusion \cite{rombach2022highresolution}, we train for 1000 time steps with conditioned diffusion model and use 50 steps of DDIM \cite{song2021denoising} at inference to sample a completed observed reflectance map. 

All networks are trained with exponential moving average of decay rate 0.9999.

The trajectory of the reflectance parameter $\Psi^{(k)}$ that dictates the forward and reverse process is determined by \cref{eq:reverse-step-base-definition}
where $\eta$ controls the rate of change towards perfect mirror reflection $\Psi_0$. 
In our experiments, we use $\eta=0.95$. 
We set $\epsilon$ which is used as a threshold to determine convergence of $\Psi^{(k)}$ to $\Psi_0$ to 0.01.
This means that the maximum time step is $K=108$.
The variance of the additive Gaussians for the forward and reverse steps are set to $\sigma=0.02$ and $\delta=0.025$. 

\section{Dataset}

We create a large-scale synthetic reflectance map dataset. We use the Laval Indoor Dataset \cite{bolduc2023beyond, gardner2017learning} and the Poly Haven HDRIs \cite{polyhaven} as the illumination. 
We split each of these HDR environment map datasets $8:2$ into training and test sets and combine them to obtain the overall training and test sets. Every time we sample an illumination from the training set, we sample a random reflectance parameter $\Psi^{(K)}$, viewing direction, and time step $k$ and render three reflectance maps $L_r^{(K)}, L_r^{(k)},$ and $L_r^{(k-1)}$ corresponding to $\Psi^{(K)}, \Psi^{(k)},$ and $\Psi^{(k-1)}$, respectively.
The reflectance parameter $\Psi^{(K)}$ is uniformly sampled $\mathbb{R}^6; \mathbb{R}\in[0, 1]$. The view direction is sampled from a uniformly discretized set of 64 angles spanning 360 degrees, and the time step $k$ is uniformly sampled within the range of $||\Psi^{(0)} - \Psi_0||_2 < \epsilon$.
\cref{fig:training-samples} shows samples of these synthetic complete observed reflectance maps.
We used 1730 environment maps for training and randomly sampled reflectance parameters and viewpoints. The resulting number of reflectance maps used for training is about 1.7 million. Training took about 5 days with an NVIDIA A100 GPU.

These synthetic complete observed reflectance maps are also used to train ObsNet. 
We compute normal maps of the random shapes in \cite{xu2018deep} to obtain visibility masks of the reflectance map. By adding Gaussian noise to the rendered observed reflectance map and then by masking it with this visibility mask, we obtain sparse observed reflectance maps, which are paired with its complete original to train ObsNet.
We also fine-tune it with raw reflectance maps from synthetically rendered random shapes for robustness against global illumination.
As all reflectance maps are in HDR, we apply log-scale transformations to pass them through each network. For ObsNet, we take the logarithm for each reflectance map independently and linearly map them to $[-1, 1]$.
For DRMNet, we normalize the overall scale of the forward and reverse process based on the intensity of the observed reflectance map and compress the brighter values with $\log_{10}(x + 0.1) + 1$. The Gaussian noise for each model is applied after these intensity transforms.

\section{Implementations of Past Methods}

In this section, we elaborate on the prerequisites and implementations of previous methods for comparative experimental evaluation.

\vspace{-8pt}
\paragraph{LIME \cite{meka2018lime}}
is a method for estimating homogeneous reflectance and an environment map.
While this method does not use the object geometry to estimate the reflectance, it needs a normal map corresponding to the input object image for environment map estimation.
We use the ground truth normal map for fairness. The environment map is estimated by mapping specular reflection using the normal map and adding approximate low frequency illumination with spherical harmonics up to the third order from diffuse reflection.

\vspace{-8pt}
\paragraph{DPI \cite{lyu2023dpi}}
estimates a spatial-varying BRDF and environment map from images.
For fair comparison, we constrain the BRDF to be homogeneous. Otherwise, the surrounding environment reflected on the object surface would be baked into the spatial-varying BRDF and the estimated environment maps become random.

\vspace{-8pt}
\paragraph{ALP \cite{yu2023accidental}}
needs to first compute a spatial-varying BRDF from multiple images of an object with known illumination and geometry. Only after that, the method can estimate an environment map from an image of the same object placed in a different environment.
To compare this method with ours on the nLMVS-Real dataset, we use the multi-view images in the ``laboratory'' environment to pre-acquire the BRDF. 
We use this pre-acquired BRDF to run ALP on images in other environments, \ie, ``buildings/chapel/court/entrance/manor.''

\vspace{-8pt}
\paragraph{Zhu \etal \cite{zhu2022learning}}
estimate spatial-varying BRDF, geometry, and out-of-view illumination from a single image of complex indoor scenes.
We obtain complete environment maps by estimating the out-of-view area at the center position of the input image from the network.
This method is significantly different from ours in its assumptions (wide field-of-view input images), so we only compare it within this supplemental material.

\section{Additional Qualitative Results}
\begin{figure*}[t]
    \centering
    \includegraphics[width=\linewidth]{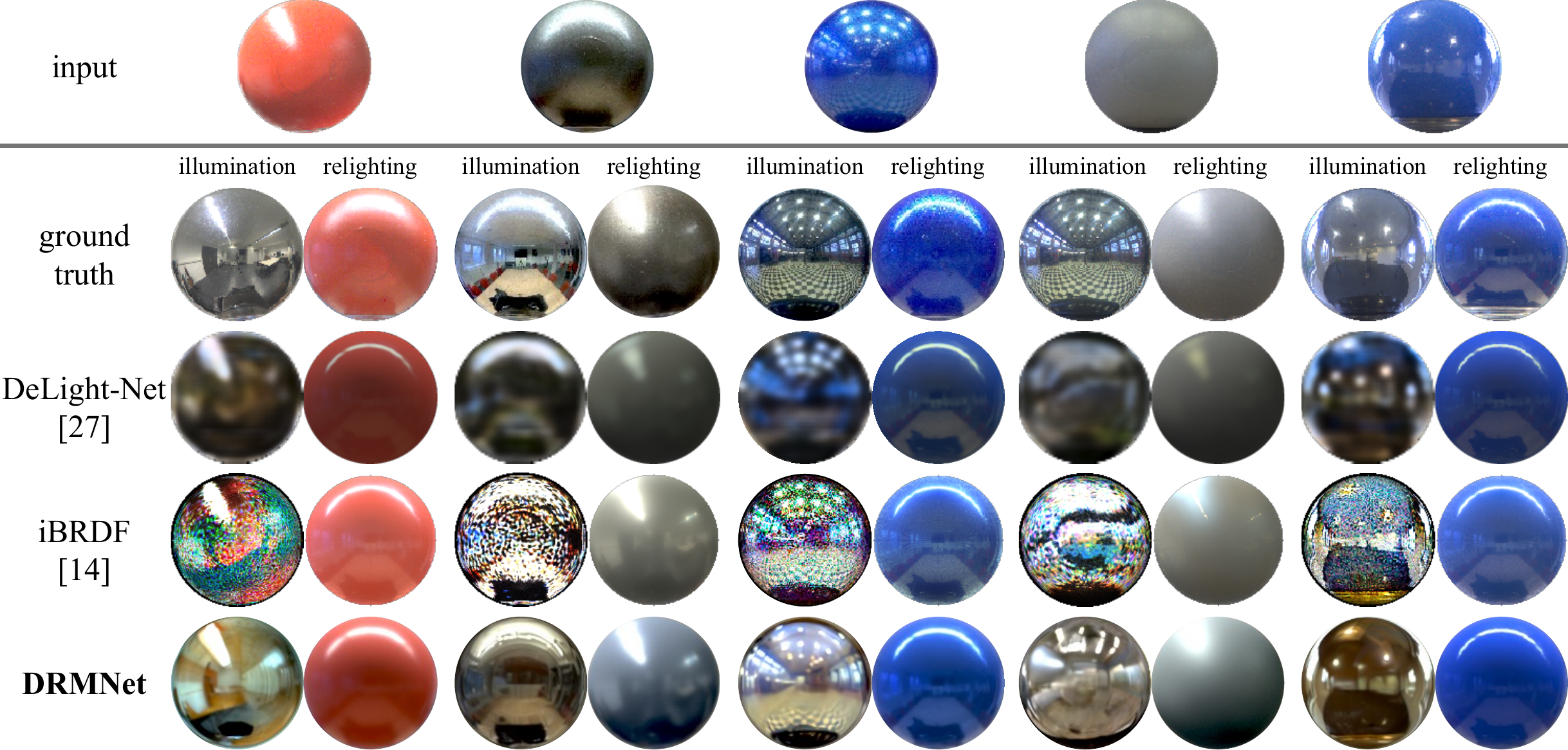}
    \caption{Qualitative results for the DeLight-Net dataset set \cite{georgoulis2018reflectance}. In comparison with iBRDF \cite{chen2022invertible} and DeLight-Net \cite{georgoulis2018reflectance}, DRMNet achieves qualitatively natural estimation for illumination and reflectance.}
    \label{fig:delightnet}
\end{figure*}

\Cref{fig:additional-nlmvs-relight} and \cref{fig:additional-nlmvs-objreplace} show additional relighting and object replacement results for the nLMVS-Real dataset \cite{yamashita2023nlmvs}. We also compare with Zhu \etal \cite{zhu2022learning} as it explicitly recovers the BRDF and illumination in the course of indoor inverse-rendering.
As it estimates spatial-varying BRDF, the results of each method in \cref{fig:additional-nlmvs-relight} show the objects with the same orientation as the inputs. On the other hand, the objects in the last column labeled as ``reference'' have a different orientation, because the viewpoint varies across environments in the nLMVS-Real dataset.
\Cref{fig:delightnet} shows the quantitative results on the Delight-Net dataset set \cite{georgoulis2018reflectance}. Our results are qualitatively more accurate, cleanly recovering the missing frequency components of the illumination, also evident in the object replacement results, while attaining more accuracy reflectance close to the ground truth relighting compared with other methods which reconstruct arbitrary frequency characteristics of the illumination. iBRDF \cite{chen2022invertible} comes close in quantitative accuracy but the high-frequency components of the illumination tends to be overestimated as evident in the object replacement results. Note again that ALP knows the reflectance.

\section{Ablation Study}

We validate the architecture of DRMNet by ablating its components and comparing them with the full model. We consider three sets of ablation studies. 
``W/o $\Psi^{(k)}$'' eliminates the step-wise reflectance estimate as input to IllNet so that the illumination and reflectance estimation are achieved independently. This ablation studies the importance of the confluence of jointly estimating the reflectance and the illumination, rather than independently. ``W/o $L_{r}^{(K)}$'' eliminates the conditioning on the observed reflectance map $L^{(K)}$ and achieves the iterative inversion solely based on the previous reflectance map estimate. This ablation studies the importance of referring to the observed reflectance map at every step of illumination estimation.  ``Once'' estimates the reflectance from the observed reflectance map $L_r^{(K)}$ and reuses this initial estimate in the recursive diffusion process. 

\begin{figure*}[t]
    \centering
    \includegraphics[width=\linewidth]{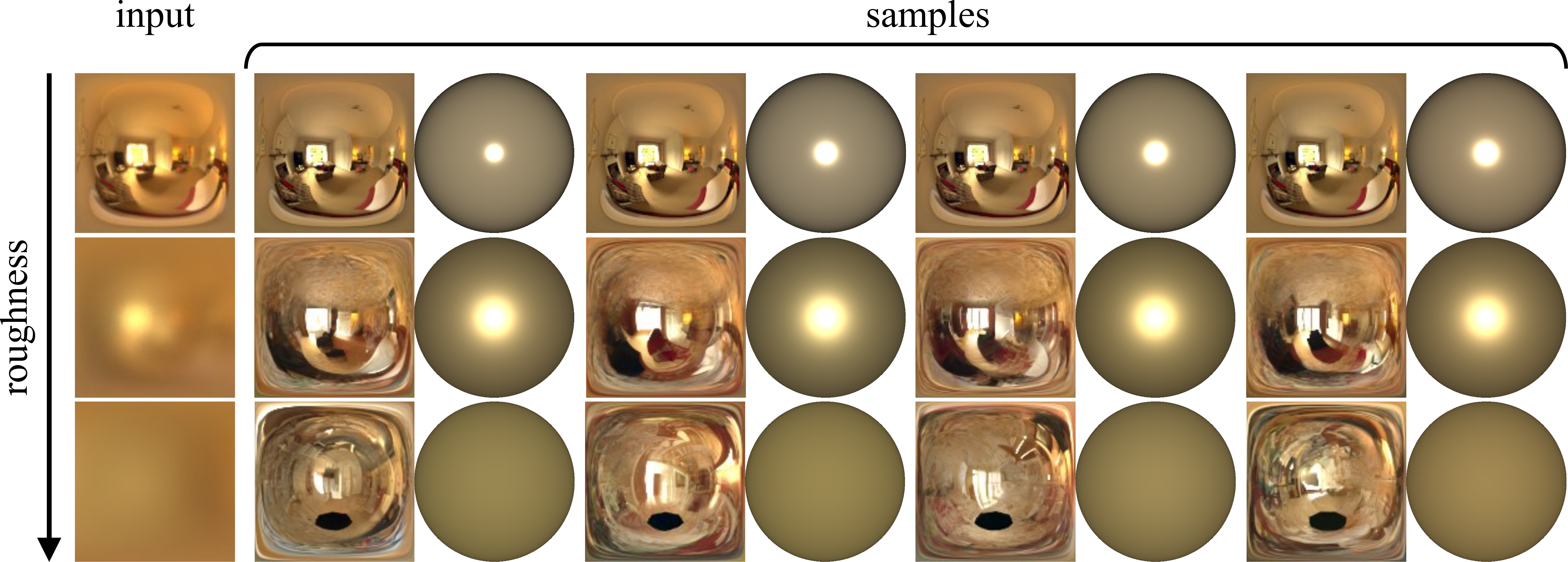}
    \caption{Results of multiple runs of DRMNet on the same input image. The left column shows the observed reflectance maps, and the right columns show the estimated samples of illumination (left) and reflectance (right) for each observation. See text for details.}
    \label{fig:multi-estimation}
\end{figure*}

\Cref{tab:ablation} shows quantitative results. The results clearly show that conditioning IllNet on the observed reflectance map $L_r^{(K)}$ which explicitly embodies the forward radiometric forward process is essential, and that the illumination and reflectance estimation processes are intertwined so that conditioning IllNet on the current estimate and recursively refining the reflectance estimate itself leads to more accurate estimates of both. This is likely because the estimation in the DRMNet recursion operates like alternating optimization leading to stable and consistent estimation. The conditioning on $L_r^{(K)}$ helps the recursive estimation of the illumination and reflectance to remain consistent with the object appearance when combined acting like a reconstruction loss. 

In contrast, the ablation result on the reflectance estimation ``once'' is counter-intuitive and we find it to be inconclusive. Estimation of the reflectance in one-shot leads to higher accuracy in logRMSE of the reflectance estimate. The reason why we still employ the iterative reflectance estimation is that we empirically found that the training and inference were more robust with this choice (also seen in the slight drop in logRMSE of illumination estimates), especially for real data. We believe this discrepancy from intuition and empirical test, manifesting particularly in logRMSE also reflects the difficulty of evaluating the ``goodness'' of a network for a generative task. We plan to further study this in more detail.

\begin{table}[t]
    \centering
    \footnotesize
    \begin{tabular}{lcccc}
                                            & \multicolumn{3}{c}{illumination} & \multicolumn{1}{c}{reflectance}         \\\cline{2-4}
                                            & logRMSE\textdownarrow                         & SSIM\textuparrow                            & LPIPS\textdownarrow      & logRMSE\textdownarrow    \\\hline
        w/o $\Psi^{(k)}$                    & 2.87                             & 0.42                            & 0.54       & 0.29        \\
        w/o $L_\mathrm{r}^{(K)}$            & 2.50                             & 0.40                            & 0.57       & \underline{0.25}        \\
        once                                & \underline{2.45}                             & \textbf{0.46}                            & \textbf{0.51}       & \textbf{0.21}           \\
        full model                          & \textbf{2.41}                             & \textbf{0.46}                            & \textbf{0.51}       & \underline{0.25}        \\\hline
    \end{tabular}
    \caption{Ablation studies of DRMNet. The full model achieves the highest accuracy, confirming the importance of principled joint estimation of illumination and reflectance with reference to the observed reflectance map realized through interdependent conditioning within DRMNet. }
    \label{tab:ablation}
\end{table}

\section{Stochastic Behavior}

We analyze the stochastic variability of our method. DRMNet seamlessly integrates stochasticity in the inverse rendering process via a reverse diffusion process on the additive Gaussian observation noise of radiometric image formation. This enables estimation of illumination faithful to the observation with stochastic variability without separate sampling. \Cref{fig:multi-estimation} shows the results of estimating the illumination and reflectance multiple times for the same input images from a set of input images under different illumination and of objects with different surface roughnesses. For the same observed reflectance map, a variation of illumination environments are estimated and their variance is large for dull reflectance closer to Lambertian and decreases for more specular reflectance centered around the ground truth. The larger the surface roughness, the wider-band of high-frequency of illumination are attenuated which is accurately reflected in these results. Note how well the recovered reflectance maps preserve the overall structure of the illumination up to the necessary frequencies---it respects the observation as much as it needs to. This is in sharp contrast to other methods that completely hallucinate an environment from noise. The reflectance estimates vary accordingly which are consistent with the observed reflectance map when combined with the corresponding illumination estimates. These results clearly show that DRMNet canonically solves stochastic inverse rendering while capturing the ambiguity between the illumination and reflectance.


%% file: figs/nlmvs-real_relighting_additional_sep.tex
\begingroup%
\makeatletter%
\providecommand\color[2][]{%
  \errmessage{(Inkscape) Color is used for the text in Inkscape, but the package 'color.sty' is not loaded}%
  \renewcommand\color[2][]{}%
}%
\providecommand\transparent[1]{%
  \errmessage{(Inkscape) Transparency is used (non-zero) for the text in Inkscape, but the package 'transparent.sty' is not loaded}%
  \renewcommand\transparent[1]{}%
}%
\providecommand\rotatebox[2]{#2}%
\newcommand*\fsize{\dimexpr\f@size pt\relax}%
\newcommand*\lineheight[1]{\fontsize{\fsize}{#1\fsize}\selectfont}%
\ifx\svgwidth\undefined%
  \setlength{\unitlength}{2837.25073242bp}%
  \ifx\svgscale\undefined%
    \relax%
  \else%
    \setlength{\unitlength}{\unitlength * \real{\svgscale}}%
  \fi%
\else%
  \setlength{\unitlength}{\svgwidth}%
\fi%
\global\let\svgwidth\undefined%
\global\let\svgscale\undefined%
\makeatother%
\begin{small}
  \begin{picture}(1,0.84686366)%
    \lineheight{1}%
    \setlength\tabcolsep{0pt}%
    \put(0.0564824,0.83346516){\makebox(0,0)[lt]{\lineheight{1.25}\smash{\begin{tabular}[t]{l}input\end{tabular}}}}%
    \put(0.17183659,0.83346516){\makebox(0,0)[lt]{\lineheight{1.25}\smash{\begin{tabular}[t]{l}Zhu et al. \cite{zhu2022learning}\end{tabular}}}}%
    \put(0.3269859,0.83346516){\makebox(0,0)[lt]{\lineheight{1.25}\smash{\begin{tabular}[t]{l}LIME \cite{meka2018lime}\end{tabular}}}}%
    \put(0.46238687,0.83346516){\makebox(0,0)[lt]{\lineheight{1.25}\smash{\begin{tabular}[t]{l}iBRDF \cite{chen2022invertible}\end{tabular}}}}%
    \put(0.6145835,0.83346516){\makebox(0,0)[lt]{\lineheight{1.25}\smash{\begin{tabular}[t]{l}DPI \cite{lyu2023dpi}\end{tabular}}}}%
    \put(0.75481608,0.83346516){\makebox(0,0)[lt]{\lineheight{1.25}\smash{\begin{tabular}[t]{l}\textbf{DRMNet}\end{tabular}}}}%
    \put(0.89413405,0.83346516){\makebox(0,0)[lt]{\lineheight{1.25}\smash{\begin{tabular}[t]{l}reference\end{tabular}}}}%
    \put(0,0){\includegraphics[width=\unitlength,page=1]{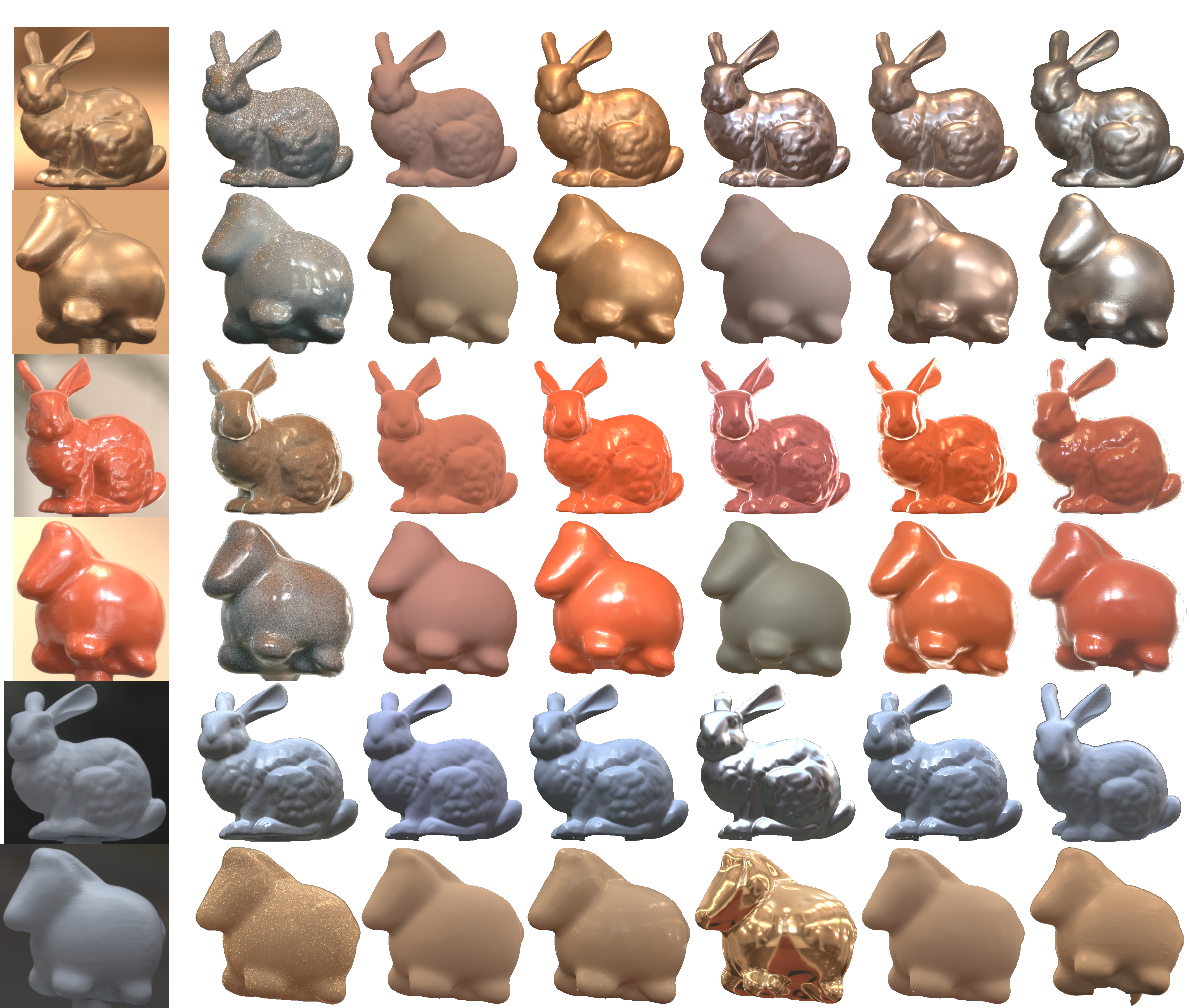}}%
  \end{picture}%
\end{small}
\endgroup%

%% file: figs/nlmvs-real_objreplace_additional_sep.tex
\begingroup%
\makeatletter%
\providecommand\color[2][]{%
  \errmessage{(Inkscape) Color is used for the text in Inkscape, but the package 'color.sty' is not loaded}%
  \renewcommand\color[2][]{}%
}%
\providecommand\transparent[1]{%
  \errmessage{(Inkscape) Transparency is used (non-zero) for the text in Inkscape, but the package 'transparent.sty' is not loaded}%
  \renewcommand\transparent[1]{}%
}%
\providecommand\rotatebox[2]{#2}%
\newcommand*\fsize{\dimexpr\f@size pt\relax}%
\newcommand*\lineheight[1]{\fontsize{\fsize}{#1\fsize}\selectfont}%
\ifx\svgwidth\undefined%
  \setlength{\unitlength}{3505.50036621bp}%
  \ifx\svgscale\undefined%
    \relax%
  \else%
    \setlength{\unitlength}{\unitlength * \real{\svgscale}}%
  \fi%
\else%
  \setlength{\unitlength}{\svgwidth}%
\fi%
\global\let\svgwidth\undefined%
\global\let\svgscale\undefined%
\makeatother%
\begin{small}
  \begin{picture}(1,0.68798194)%
    \lineheight{1}%
    \setlength\tabcolsep{0pt}%
    \put(0.06099586,0.67565205){\makebox(0,0)[lt]{\lineheight{1.25}\smash{\begin{tabular}[t]{l}input\end{tabular}}}}%
    \put(0.15506994,0.67565205){\makebox(0,0)[lt]{\lineheight{1.25}\smash{\begin{tabular}[t]{l}Zhu \etal \cite{zhu2022learning}\end{tabular}}}}%
    \put(0.29646338,0.67565205){\makebox(0,0)[lt]{\lineheight{1.25}\smash{\begin{tabular}[t]{l}LIME \cite{meka2018lime}\end{tabular}}}}%
    \put(0.41174792,0.67565205){\makebox(0,0)[lt]{\lineheight{1.25}\smash{\begin{tabular}[t]{l}iBRDF \cite{chen2022invertible}\end{tabular}}}}%
    \put(0.54152326,0.67565205){\makebox(0,0)[lt]{\lineheight{1.25}\smash{\begin{tabular}[t]{l}DPI \cite{lyu2023dpi}\end{tabular}}}}%
    \put(0.66731058,0.67565205){\makebox(0,0)[lt]{\lineheight{1.25}\smash{\begin{tabular}[t]{l}ALP \cite{yu2023accidental}\end{tabular}}}}%
    \put(0.78796098,0.67565205){\makebox(0,0)[lt]{\lineheight{1.25}\smash{\begin{tabular}[t]{l}\textbf{DRMNet}\end{tabular}}}}%
    \put(0.92938587,0.67565205){\makebox(0,0)[lt]{\lineheight{1.25}\smash{\begin{tabular}[t]{l}GT\end{tabular}}}}%
    \put(0,0){\includegraphics[width=\unitlength,page=1]{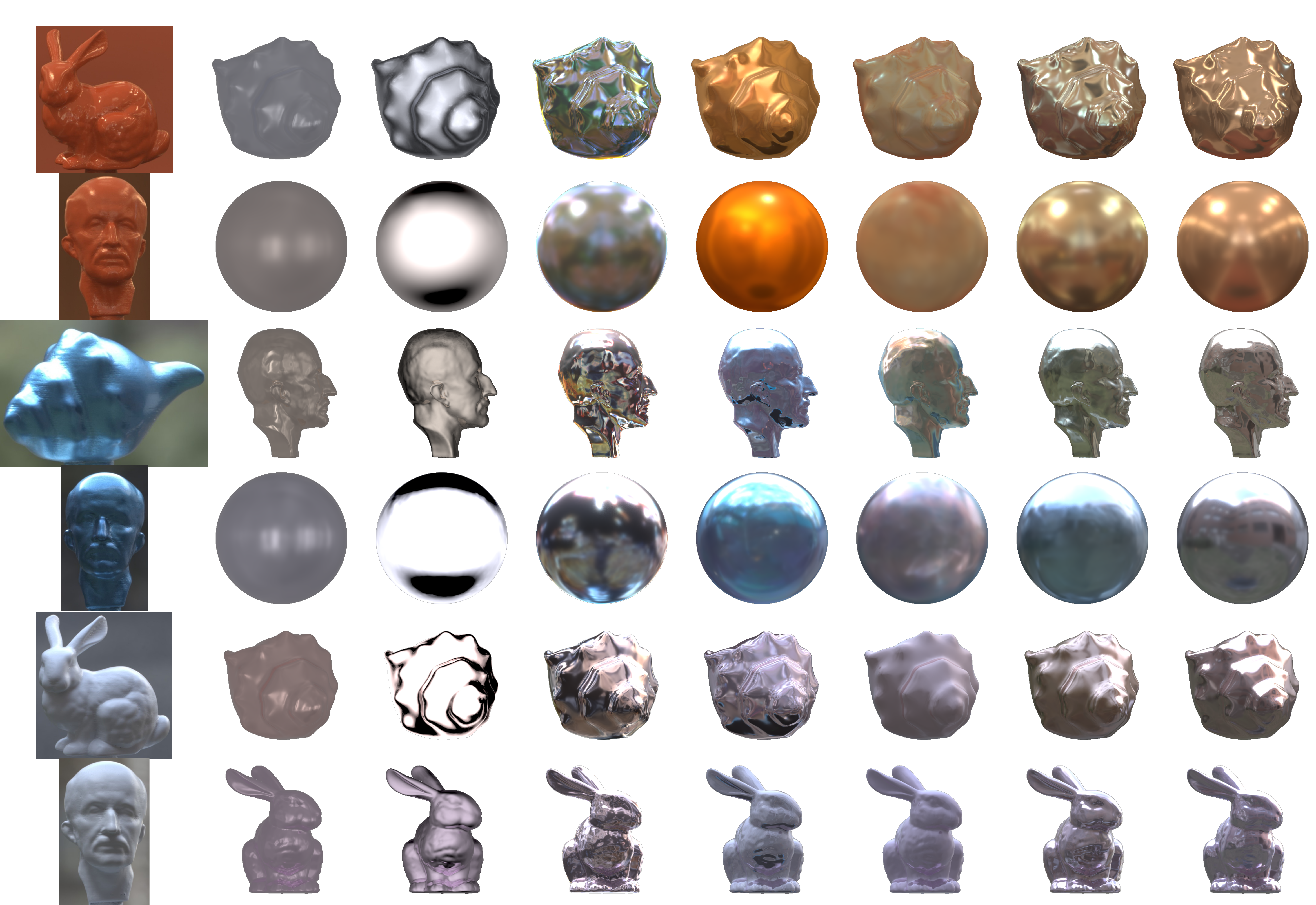}}%
  \end{picture}%
\end{small}
\endgroup%